\documentclass[sn-apa]{sn-jnl}

\usepackage{graphicx}
\usepackage{multirow}
\usepackage{amsmath,amssymb,amsfonts}
\usepackage{amsthm}
\usepackage{mathrsfs}
\usepackage[title]{appendix}
\usepackage{xcolor}
\usepackage{textcomp}
\usepackage{manyfoot}
\usepackage{booktabs}
\usepackage{algorithm}
\usepackage{algorithmicx}
\usepackage{algpseudocode}
\usepackage{listings}
\usepackage{pdflscape}
\usepackage{makecell}

\begin{document}

\title[Anomalous Change Point Detection Using Probabilistic Predictive Coding]{Anomalous Change Point Detection Using Probabilistic Predictive Coding}

\author[1]{\fnm{Roelof G.} \sur{Hup}}\email{r.g.hup@tue.nl} 
\author*[1]{\fnm{Julian P.} \sur{Merkofer}}\email{j.p.merkofer@tue.nl} 
\author[2]{\fnm{Alex A.} \sur{Bhogal}}\email{a.bhogal@umcutrecht.nl} 
\author[1]{\fnm{Ruud J.G.} \sur{van Sloun}}\email{r.j.g.v.sloun@tue.nl} 
\author[3]{\fnm{Reinder} \sur{Haakma}}\email{reinder.haakma@philips.com} 
\author[1]{\fnm{Rik} \sur{Vullings}}\email{r.vullings@tue.nl} 

\affil[1]{\orgdiv{Department of Electrical Engineering}, \orgname{Eindhoven University of Technology}, \orgaddress{\city{Eindhoven}, \country{The Netherlands}}}
\affil[2]{\orgdiv{Translational Neuroimaging Group, Center for Image Sciences},  \orgname{University Medical Center Utrecht}, \orgaddress{\state{Utrecht}, \country{The Netherlands}}}
\affil[3]{\orgdiv{Department of Patient Care \& Monitoring, Philips Research}, \orgname{Koninklijke Philips N.V.}, \orgaddress{\city{Eindhoven}, \country{The Netherlands}}}

\abstract{Change point detection (CPD) and anomaly detection (AD) are essential techniques in various fields to identify abrupt changes or abnormal data instances. However, existing methods are often constrained to univariate data, face scalability challenges with large datasets due to computational demands, and experience reduced performance with high-dimensional or intricate data, as well as hidden anomalies. Furthermore, they often lack interpretability and adaptability to domain-specific knowledge, which limits their versatility across different fields. In this work, we propose a deep learning-based CPD/AD method called Probabilistic Predictive Coding (PPC) that jointly learns to encode sequential data to low-dimensional latent space representations and to predict the subsequent data representations as well as the corresponding prediction uncertainties. The model parameters are optimized with maximum likelihood estimation by comparing these predictions with the true encodings. At the time of application, the true and predicted encodings are used to determine the probability of conformance, an interpretable and meaningful anomaly score. Furthermore, our approach has linear time complexity, scalability issues are prevented, and the method can easily be adjusted to a wide range of data types and intricate applications. We demonstrate the effectiveness and adaptability of our proposed method across synthetic time series experiments, image data, and real-world magnetic resonance spectroscopic imaging data.}

\keywords{change point detection, anomaly detection, predictability modeling, deep learning}

\maketitle

\section{Introduction}
Change point detection (CPD) and anomaly detection (AD) are critical techniques in various fields, involving the detection of abrupt changes or abnormal instances in sequential data, respectively. Both techniques are widely used in financial surveillance and fraud detection \citep{Hilal2022FinancialAdvances}, Internet of Things and (remote) health monitoring \citep{Fahim2019AnomalyReview}, video surveillance \citep{Duong2023DeepSurvey}, and intrusion detection in cybersecurity systems \citep{Yang2022ADetection}. With ever-increasing amounts of data, these techniques have become more important, as they decrease labor efforts and enable the detection of subtle but significant anomalies and change points.

When the objective is to detect anomalous change points, the fields of CPD and AD intersect, in the sense that change points can be either normal or anomalous. For example, in the field of computer vision and video surveillance, changes in image contrast or brightness are generally not anomalous, while substantial changes may be anomalous \citep{Theiler2006ProposedDetection}. Similarly, in fields like water resource management \citep{Apostol2021ChangeData} and hardware Trojan detection \citep{Elnaggar2019HardwareTechniques}, failing to consider the existence of normal change points in anomaly detection leads to numerous false alarms, as routine changes may trigger unnecessary alerts.

Despite the importance of CPD and AD, existing methods often encounter significant limitations. These include constraints on data types, such as being applicable only to univariate data, scalability issues with large datasets due to high computational complexity, and performance degradation with high-dimensional, complex data, or concealed anomalies. Moreover, many methods lack interpretability, fail to provide meaningful scores, cannot be tuned to domain knowledge, or are overly specialized in particular domains, hindering their cross-domain applicability. These limitations are further discussed in section \ref{section:background}.

In this work, we propose Probabilistic Predictive Coding (PPC), a method that does not suffer from these limitations. PPC is a deep learning-based CPD/AD method that encodes sequential data to latent space representations and thus can be adapted to a wide range of data types, including time series, images, videos, graphs, sets, and text data. Due to the application of deep learning, the computational complexity at application time is linear with respect to the number of samples, preventing scalability issues. At the same time, the deep learning approach makes the method widely applicable in different domains with different levels of data complexity, while still being able to be tuned to domain knowledge. By using predictability modeling, the method is intuitive and interpretable. Furthermore, due to its probabilistic nature, the method can provide a meaningful anomaly score in the form of a \textit{probability of conformance}.

\section{Background}
\label{section:background}

\subsection{Applications of CPD and AD}\label{section:bg-applications}
In this section, we provide a non-exhaustive list of applications of change point detection and anomaly detection.
\\\\
\noindent\textbf{Financial surveillance}: In the financial sector, financial fraud detection systems for credit card fraud, insurance fraud, money laundering, healthcare fraud, and securities and commodities fraud have gained interest due to increasing incidence rates. In the US alone, yearly costs associated with financial fraud are estimated to accumulate to over \$400 billion \citep{Hilal2022FinancialAdvances}. We refer the reader to an extensive review on this topic by \cite{Hilal2022FinancialAdvances}.
\\\\
\noindent\textbf{Sensor monitoring}: Due to the advancement of sensor monitoring technologies within physical spaces and objects, e.g., inhabitant environments, transportation systems, health care systems, and industrial systems, the need for automated analysis of large data streams has become apparent. Timely detection of malfunctioning equipment may prevent unexpected problems \citep{Fahim2019AnomalyReview}. The work of \cite{Fahim2019AnomalyReview} provides a literature review on this topic.
\\\\
\noindent\textbf{Video surveillance}: With an increasing demand for public security, the use of surveillance cameras for crime prevention and counter-terrorism has risen. However, manual human monitoring is labor-intensive, especially given the low incidence rate of abnormal events. In the field of video surveillance, anomaly detection for automated monitoring is an active area of research \citep{Duong2023DeepSurvey}. We refer the reader to a survey for deep learning-based video surveillance approaches by \cite{Duong2023DeepSurvey}.
\\\\
\noindent\textbf{Cybersecurity systems}: For decades, anomaly detection systems have been used to monitor network traffic in computer systems for malicious requests. Research into intrusion detection systems (IDS) has been active, especially within the machine learning community \citep{Yang2022ADetection}. The work of \cite{Yang2022ADetection} contains a systematic literature review on this topic.

\subsection{Anomaly detection methods} \label{section:bg-ad-methods}
Anomaly detection methods aim to identify data instances that deviate significantly from the expected behavior of normal data. Over the decades, approaches have evolved from statistical hypothesis tests to probabilistic and deep learning–based methods, reflecting a shift from static modeling to dynamic, data-driven prediction. While we will mention some highlights of the developments over the years, we refer the reader to comprehensive reviews like the work of \cite{Paparrizos2025AdvancesMeasures} for in-depth information.

Early statistical approaches relied on parametric hypothesis testing and robust statistics. Grubbs’ test \citep{Grubbs1969ProceduresSamples} detects univariate outliers via Studentized range thresholds, while the multivariate $\chi^2$-AD method \citep{Ye2001AnSystems} generalizes this principle using chi-squared statistics for multivariate or network data. Such models offer interpretability and computational efficiency but depend on assumptions of normality and independence that rarely hold in complex, non-stationary settings. Early time-series models, such as ARIMA \citep{Box2016TimeControl} and Exponential Smoothing (ES, DES, TES; \citealp{Snyder1983ExponentialCorrection}), extended this paradigm to temporal prediction, but remained limited to linear dynamics and Gaussian noise.

Proximity- and density-based methods reframed anomalies as geometrically or statistically isolated points in feature spaces. Distance-Based Outlier Detection (DBOD, \citealp{Knorr2000Distance-basedApplications}) identifies points with sparse $R$-neighbourhoods, while Local Outlier Factor (LOF, \citealp{Breunig2000LOF:Outliers}) measures local reachability density relative to neighbors. Histogram-Based Outlier Detection (HBOD, \citealp{Gebski2007AnDetection}) and HDBSCAN \citep{Campello2015HierarchicalDetection} improve scalability and hierarchical robustness.

Clustering-based methods such as Two-Phase Clustering (TPC, \citealp{Jiang2001Two-phaseDetection}) and Clustering-Based Outlier Detection (CBOD, \citealp{Jiang2008Clustering-BasedMethod}) interpret anomalies as data points that do not belong to any dense cluster. While effective for low-dimensional, static data, these methods are ineffective in high-dimensional or evolving distributions.

Subspace and ensemble approaches improved scalability and robustness. The Subspace Outlier Degree (SOD, \citealp{Kriegel2009OutlierData}) identifies anomalies within relevant feature subspaces, while ensemble frameworks such as ZERO++ \citep{Pang2016ZERO++:Sets} combine feature projections for stability. Isolation Forest (iForest, \citealp{Liu2012Isolation-BasedDetection}) isolates anomalies by randomly partitioning data. However, these techniques generally ignore temporal dependencies.

Forecasting and reservoir-based sequence models emerged to handle sequential data. Echo State Networks (CoalESN, MoteESN; \citealp{Obst2008UsingMines, Chang2009Mote-BasedNetworks}) applied recurrent reservoirs for streaming anomaly detection, while Probabilistic Change Identification (PCI, \citealp{Yu2014TimePrediction}) incorporated adaptive thresholding. Yet, these methods often assume stationarity and lack principled uncertainty estimation.

Deep representation learning introduced end-to-end feature extraction. Recurrent architectures such as LSTM-AD \citep{Malhotra2015LongSeries} and EncDec-AD \citep{Malhotra2016} train encoder–decoder LSTMs on normal sequences, detecting anomalies via reconstruction error. OmniAnomaly \citep{Su2019RobustNetwork} extends this with stochastic latent variables for robustness.

Generative probabilistic models such as VAE-LSTM \citep{Lin2020AnomalyModel} model normal dynamics through probabilistic reconstruction, while GAN-based variants like f-AnoGAN \citep{Schlegl2019F-AnoGAN:Networks}, TadGAN \citep{Geiger2020TadGAN:Networks}, and TAnoGAN \citep{Bashar2020TAnoGAN:Networks} learn adversarial latent representations of normality. Hybrids such as USAD \citep{Audibert2020USAD:Series} jointly optimize reconstruction and prediction losses to tighten anomaly bounds, though thresholding remains heuristic.

Transformer-based methods have recently achieved state-of-the-art performance on multivariate and long-range dependencies. MTAD-GAT \citep{Zhao2020MultivariateNetwork} incorporates graph attention for inter-variable relations, Anomaly Transformer \citep{Xu2022} quantifies associations via self-attention discrepancy, and TranAD \citep{Tuli2022} and PatchAD \citep{Zhong2025PatchAD:Detection} use efficient patch-level encoders for scalable reconstruction. TimesNet \citep{Wu2023CLformer:Structures} reformulates 1D series as 2D temporal patterns, while AnomalyLLM \citep{Liu2024AnomalyLLM:Models} fine-tunes large language models for zero-shot anomaly reasoning.

\subsection{Change point detection methods} \label{section:bg-cpd-methods}
Change point detection (CPD) aims to identify points in time where the properties of a process change abruptly, causing significant shifts in sequential data. Like AD methods, CPD methods have evolved from simpler statistical tests to deep learning-based solutions. Highlights of developments are mentioned here, but more comprehensive surveys can be found in \cite{Truong2020SelectiveMethods} and \cite{Xu2025Change-pointReview}.

Early CPD research focused on sequential likelihood-ratio methods. The cumulative sum (CUSUM, \citealp{Page1954ContinuousSchemes}) test introduces recursive accumulation of log-likelihood differences between competing hypotheses of stationarity and change. Together with related sequential probability ratio tests \citep{Wald1945SequentialHypotheses}, CUSUM forms the foundation of online detection theory, defining optimality in terms of the average run length to false alarm. 

Later developments reinterpreted such sequential schemes through a Bayesian lens. The Bayesian Online Changepoint Detection (BOCPD, \citealp{Adams2007BayesianDetection}) algorithm expresses the posterior over the run length since the last change as a recursive message-passing update weighted by the model’s predictive likelihood. BOCPD established the modern probabilistic framing of online CPD and inspired extensions to streaming, high-dimensional, and budgeted settings, e.g., \cite{Wang2021OnlineBudget}.

Parallel to sequential approaches, research was focused on offline change point detection, where the entire data sequence is observed and optimal segmentation is desired. Early econometric contributions, such as \cite{Bai1998EstimatingChanges}, formulated CPD as a model-selection problem over linear regression segments, combining Wald-type test statistics with trimming strategies to ensure asymptotic consistency. 

Subsequent research developed exact search algorithms under penalized cost formulations. The Optimal Partitioning and Pruned Exact Linear Time (PELT, \citealp{Killick2012OptimalCost}) method achieves linear computational cost by pruning candidate boundaries while maintaining global optimality under additive penalties. Randomized algorithms such as Wild Binary Segmentation (WBS; \citealp{Fryzlewicz2014WildDetection}) improved practical sensitivity to localized or short-lived changes by evaluating CUSUM statistics over random subintervals rather than fixed partitions. Together, these methods established the dominant penalized optimization paradigm for multiple change point inference.

While parametric likelihood formulations dominate classical CPD, many real-world applications exhibit changes beyond shifts in mean or variance. Nonparametric methods address this by comparing empirical distributions across adjacent windows without assuming a specific distribution. The energy-distance approach of \cite{Matteson2014AData} detects distributional changes by maximizing between-segment dissimilarity in terms of pairwise distances, yielding consistency for multivariate and heavy-tailed data. Kernel-based detectors, notably those by \cite{Harchaoui2008KernelAnalysis}, leverage reproducing kernel Hilbert space embeddings to detect nonlinear and high-order distributional changes via maximum mean discrepancy or Hilbert–Schmidt independence criteria. These approaches generalize the notion of a contrast function to measure divergence between local empirical distributions.

Recent research has expanded CPD to structured and high-dimensional domains. The Group Fused Lasso \citep{Bleakley2011TheDetection} enforces joint sparsity across multiple correlated series, identifying shared change points via total-variation regularization. High-dimensional theoretical analyses, such as those by \cite{Cho2015Multiple-Change-PointSegmentation}, derive consistency guarantees for sparse change recovery in regimes where the number of variables exceeds the sample size. Sliding-window methods like MOSUM \citep{Chu1995MOSUMConstancy} and its multivariate extensions, e.g., \cite{McGonigle2025NonparametricFunctions}, adapt moving-sum test statistics to high-dimensional covariance structures, allowing scalable online operation. These approaches extend classical univariate tests to complex multivariate and structured data.

In parallel, modern representation-learning techniques employ learned or neural approaches to CPD. Deep architectures for time series segmentation now frame change detection as an end-to-end learning problem, where neural networks infer latent representations that linearize temporal regimes and produce differentiable change scores \citep{Li2024AutomaticLearning,Nguyen2025PenaltyPerceptron}. Other recent work explores continuous-time latent stochastic differential equation models for CPD \citep{Ryzhikov2023LatentDetection}, merging neural dynamics with Bayesian inference. Although many deep methods achieve strong performance, they typically depend on supervised labels or synthetic data generation to train segmentation boundaries.

Recent advances have also introduced unsupervised deep learning approaches for CPD, which reduce reliance on labeled change points. Adaptive LSTM–autoencoder architectures have been proposed for memory-free online CPD \citep{Atashgahi2023Memory-freeApproach}, while graph neural network encoder–decoder models capture evolving correlations in multivariate sequences \citep{Zhang2020Correlation-AwareNetworks}. Furthermore, fully unsupervised multivariate CPD using minimum-description-length guided greedy segmentation has been developed \citep{Wu2024UnsupervisedSeries}. These unsupervised methods highlight the growing emphasis on flexible, label-free detection of structural shifts in complex time series.

\subsection{Anomalous change point detection methods}
\label{section:ACPDmethods}
Anomalous change point detection (ACPD) methods not only aim to identify distributional breaks but also to determine which of those breaks are truly anomalous instead of routine regime shifts. Early works established this two-score approach. ChangeFinder \citep{Takeuchi2006ASeries} produces both outlier and change point scores within an online, likelihood-based framework, making a clear distinction between transient atypical observations and persistent distributional changes. Bayesian latent-variable models extend this idea by inferring posterior indicators for both event types. For example, ABCO (Adaptive Bayesian Changepoint \& Outlier scoring) places local outlier processes inside a state-space forecasting model and yields posterior probabilities for changepoints and outliers \citep{Wu2024TrendScoring}, while recent online Bayesian models explicitly aim to jointly detect collective anomalies and change points in streaming, detailed series \citep{Chen2025BayesianSeries}.

Other lines of work detect anomalous change points via segmentation and practical pipelines. Penalized-cost segmentation methods that separate collective (segment) anomalies from point anomalies (e.g., the linear-time CAPA family and related algorithms) provide efficient offline labelling of anomalous segments versus point events \citep{Fisch2022AAnomalies}. In applied monitoring contexts, pipelines that use CPD outputs to suppress anomaly alarms show strong empirical value. Apostol et al. propose a CPD-enhanced anomaly pipeline for IoT data that reduces false positives by treating change points as expected regime shifts unless further evidence indicates anomaly \citep{Apostol2021ChangeData}.

Recent work moves toward multivariate, non-linear, and decision-theoretic formulations. Liu et al. developed an unsupervised approach for simultaneous anomaly and change point detection under concept drift by using fluctuation-based features and persistence testing to distinguish transient anomalies from regime shifts in multivariate series \citep{Liu2023AnomalyDrift}. On the theoretical side, formulations that seek an anomalous change in one process among many provide asymptotic optimality results for sequential search strategies \citep{Didi2024AsymptoticallyModel}.

These methods show clear trade-offs: classical/online methods offer real-time processing but limited representation power \citep{Takeuchi2006ASeries}; Bayesian models provide calibrated event posteriors but have historically assumed simple dynamics or operated offline, though recent extensions enable online processing \citep{Wu2024TrendScoring, Chen2025BayesianSeries}; segmentation and pipeline methods scale well in practice but are often heuristic and non-probabilistic \citep{Fisch2022AAnomalies, Apostol2021ChangeData}; and recent approaches begin to handle multivariate, drifting data but rarely combine deep representation learning with calibrated, change-point-specific anomaly scoring \citep{Liu2023AnomalyDrift, Didi2024AsymptoticallyModel}.

Probabilistic Predictive Coding addresses this gap by combining learned predictive latent representations with explicit probabilistic conformance scoring for change points: it retains probabilistic modeling while providing the representation power and (online/offline) applicability needed for modern multivariate sensors, thereby enabling distinction between anomalous and routine change points.

\section{Methods}

\subsection{Problem definition}
Let $\{x_1, x_2, \cdots\}$ be a set consisting of data instances of any type that are \textit{sequential}, \textit{predictable}, and \textit{encodable}. Data instances are sequential whenever they are part of a unidirectional succession, e.g., $x_2$ succeeds $x_1$, $x_3$ succeeds $x_2$, etc. The property of predictability is applicable whenever individual data instances can be reasonably estimated given prior data instances in the sequence, indicated by a relatively low entropy of the conditional probability density function $f\left(x_{j+i}|x_1, x_2, \cdots, x_j\right)$ for prediction step $i$. Data instances are encodable whenever all data instances $\{x_1, x_2, \cdots\}$ can be transformed to low-dimensional latent space representations $\{z_1, z_2, \cdots\}$ with an encoder neural network, and these latent space representations contain the necessary information to satisfy the requirement of predictability. Obvious examples of data that meet these requirements are video (e.g., with convolutional neural networks) or text (e.g., with Word2vec), but certain time series, images, and even graphs may qualify as well. It is up to the user to define a sequential axis within the data such that the data instances indeed have these properties.

Based on these properties, we propose a method that is able to encode sequential data instances into a latent space representation and make predictions of the next latent space representations in the sequence. In this way, highly intricate data can be reduced in complexity and predicted representations can be compared to the actual representations, giving us a proxy of the likelihood function $f\left(x_{j+i}|x_1, x_2, \cdots, x_j\right)$ and therefore a measure to determine whether the actual data instance was expected or likely. Within the taxonomy proposed by \cite{Pang2022DeepDetection}, this method would be considered as \textit{predictability modeling}, where the latent space representations of non-anomalous data instances are learned in a semi-supervised manner.

\subsection{Probabilistic predictions} \label{section:predictions}
The objective of the proposed method is to encode a set of $N_p$ data instances $\{x_1, x_2, \cdots, x_{N_p}\}$ to their corresponding latent space representations $\{z_1, z_2, \cdots, z_{N_p}\}$ and use these to predict the distribution(s) of the subsequent representation(s) $\{\hat{z}_{N_p+1}, \hat{z}_{N_p+2}, \cdots\}$ and the corresponding prediction uncertainty $\{\sigma_{N_p+1}, \sigma_{N_p+2}, \cdots\}$. Without loss of generality, we let $z_{N_p+i}, \hat{z}_{N_p+i}\in\mathbb{R}^{N_e}, \sigma_{N_p+i}\in\mathbb{R}^{N_e}_{>0}$ for prediction step $i$ and assume that the prediction error is distributed according to a multivariate normal distribution with zero covariance, i.e., $Z_{N_p+i}\sim\mathcal{N}\left(\hat{z}_{N_p+i}, \textrm{diag}\left(\sigma^{\circ2}_{N_p+i}\right)\right)$. Note that other continuous and differentiable probability distributions could be used as well. Discrete probability distributions could potentially be used using the Gumbel-Softmax reparameterization trick \citep{Jang2017CategoricalGumbel-Softmax,Huijben2023ALearning}.

Now, to encode the data instances and predict latent space representations of the next data instances in the sequence, we adapted the Contrastive Predictive Coding pipeline (CPC) proposed by \cite{VanDenOord2018}. While CPC is originally used for representation learning, the authors demonstrated its ability to learn high-quality representations combined with its predictive abilities, which forms a suitable basis for the objectives within the current work.

\begin{figure}[h]%
\centering
\includegraphics[width=1.0\textwidth]{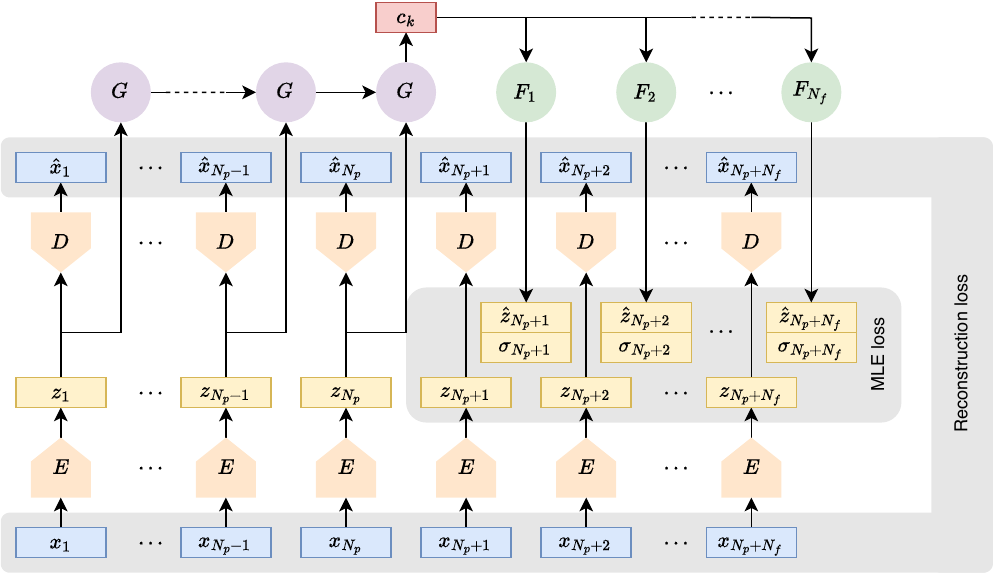}
\caption{The PPC pipeline architecture. Encoder model $E$ encodes data instances to their latent space representations, and decoder model $D$ aims to reconstruct the original data instances. Recurrent neural network model $G$ and forecasting models $F_i$ make predictions of the latent space encodings of future latent space representations given the past data instances. Maximum likelihood estimation (MLE) loss and reconstruction loss are depicted in gray.}\label{fig:PPC}
\end{figure}

An overview of the proposed PPC pipeline is shown in Figure~\ref{fig:PPC}. The architecture consists of the following key modules. The model choice for each module is flexible and can be replaced with any suitable alternative.

\begin{itemize}
    \item \textbf{Encoder ($E$):} A neural network that transforms each input data instance $x_k$ into a low-dimensional latent representation $z_k = E(x_k)$. This step compresses the input while preserving relevant features for downstream prediction. The encoder is typically implemented as a convolutional or feedforward network, depending on the data modality.

    \item \textbf{Sequence Model ($G$):} A recurrent neural network (e.g., GRU or LSTM) that processes the sequence of latent representations $\{z_1, z_2, \cdots, z_{N_p}\}$ and summarizes them into a context vector $c_{N_p} = G(z_1, z_2, \cdots, z_{N_p})$. This context vector captures temporal dependencies and serves as a compact summary of the observed sequence.

    \item \textbf{Forecasting Networks ($F_i$):} A set of neural networks, each responsible for predicting a future latent representation $\hat{z}_{N_p+i}$ and its associated uncertainty $\sigma_{N_p+i}$, i.e., $(\hat{z}_{N_p+i}, \sigma_{N_p+i}) = F_i(c_{N_p})$. These Mean Variance Estimation neural networks are trained to forecast multivariate normal distributions of the future latent representations with diagonal covariances. The approach of \citet{Nix1994EstimatingDistribution} is followed, allowing multi-task learning of both prediction and uncertainty.

    \item \textbf{Decoder ($D$):} This module plays a vital role during the training phase but is not used during the anomaly detection inference phase. Its function is to reconstruct the original input data from its latent space representation. This reconstruction acts as a regularisation mechanism during joint training, compelling the Encoder to learn meaningful and invertible latent representations. 
\end{itemize}

We opted for this probabilistic prediction framework because: (1) this allows for calculating a proxy of the likelihood $f\left(x_{N_p+i}|x_1, x_2, \cdots, x_{N_p}\right)$ (see section \ref{section:proxy}), (2) this likelihood is an intuitive and non-arbitrary distance metric between ground truth and predicted encodings and forms a suitable loss function for training neural networks (see section \ref{section:trainproc}), and (3) probabilistic predictions allow for a nuanced and statistically meaningful interpretation of the output in the form of a \textit{probability of conformance} (see section \ref{section:probconf}). Figure~\ref{fig:PPC} shows an example of the PPC pipeline architecture.

\subsection{Tractable proxy for the density function}
\label{section:proxy}
While the computation of $f\left(x_{N_p+i}|x_1, x_2, \cdots, x_{N_p}\right)$ can be intractable in highly intricate and/or high-dimensional data, given the steps taken in the previous section, a proxy can be defined for this conditional probability density function. By applying maximum likelihood optimization to the pipeline, the encoding mechanism of model $E$ and the predictive mechanism of models $G$ and $F_1, F_2, \cdots, F_{N_f}$ are jointly and symbiotically optimized to create an ordered latent space, such that models $G$ and $F_1, F_2, \cdots, F_{N_f}$, given the latent space representations within this ordered latent space, yield accurate density estimates of future latent space representations $z_{N_p+i}$.

Whenever the models accurately capture the underlying conditional distribution $f\left(x_{N_p+i}|x_1, x_2, \cdots, x_{N_p}\right)$, this distribution should be reflected in the density estimates of $z_{N_p+i}$. As the data space and latent space generally differ in volume, we assume the relationship between these two density functions to be proportional, e.g., 
\begin{align}
f\left(x_{N_p+i}|x_1, x_2, \cdots, x_{N_p}\right) &\propto f\left(E\left(x_{N_p+i}\right)|F_i\left(G\left(E\left(x_1\right), E\left(x_2\right), \cdots, E\left(x_{N_p}\right)\right)\right)\right) \nonumber \\
&= f\left(z_{N_p+i}|F_i\left(G\left(z_1, z_2, \cdots, z_{N_p}\right)\right)\right) \nonumber \\
&= f\left(z_{N_p+i}|\hat{z}_{N_p+i},\sigma_{N_p+i}\right).
\end{align}

In section \ref{section:propexp}, we will present an experiment for which this relationship holds, indicating that the pipeline indeed has the potential to estimate $f\left(z_{N_p+i}|\hat{z}_{N_p+i},\sigma_{N_p+i}\right)$ as a suitable proxy for $f\left(x_{N_p+i}|x_1, x_2, \cdots, x_{N_p}\right)$.

Furthermore, as $f\left(z_{N_p+i}|\hat{z}_{N_p+i},\sigma_{N_p+i}\right)$ takes the form of the density function of a multivariate normal distribution, i.e.,
\begin{align}
f\left(z_{N_p+i}|\hat{z}_{N_p+i},\sigma_{N_p+i}\right) = \prod_{n=1}^{N_e} \frac{1}{\sqrt{2\pi}\cdot\left(\sigma_{N_p+i}\right)_{n}}\exp\left(-\frac{1}{2}\cdot\frac{\left[\left(z_{N_p+i}\right)_n-\left(\hat{z}_{N_p+i}\right)_n\right]^2}{\left(\sigma_{N_p+i}\right)^2_n}\right),
\end{align}
it becomes clear that this proxy is tractable for both model training and application. Note that, for notation purposes, $\left(\cdot\right)_n$ denotes the $n$th element in the vector.

\subsection{Training procedure}
\label{section:trainproc}
While CPC uses InfoNCE loss \citep{VanDenOord2018}, this contrastive loss function is not compatible with the probabilistic outputs of the forecasting models. Therefore, we opted for training the models $E$, $G$ and $F_1, F_2, \cdots, F_{N_f}$ with maximum log-likelihood estimation, yielding
\begin{align}
\mathcal{L}_{\text{MLE}}&=-\frac{1}{N_f}\sum^{N_f}_{i=1}\log f\left(z_{N_p+i}|\hat{z}_{N_p+i},\sigma_{N_p+i}\right)\nonumber\\
&=-\frac{1}{N_f}\sum^{N_f}_{i=1}\log \prod_{n=1}^{N_e} \frac{1}{\sqrt{2\pi}\cdot\left(\sigma_{N_p+i}\right)_{n}}\exp\left(-\frac{1}{2}\cdot\frac{\left[\left(z_{N_p+i}\right)_n-\left(\hat{z}_{N_p+i}\right)_n\right]^2}{\left(\sigma_{N_p+i}\right)^2_n}\right)\nonumber\\
&=N_e\log\left(\sqrt{2\pi}\right)+\frac{1}{N_f}\sum^{N_f}_{i=1}\left(\sum_{n=1}^{N_e}\log\left(\sigma_{N_p+i}\right)_n+\frac{1}{2}\sum_{n=1}^{N_e}\frac{\left[\left(z_{N_p+i}\right)_n-\left(\hat{z}_{N_p+i}\right)_n\right]^2}{\left(\sigma_{N_p+i}\right)^2_n}\right),
\label{eq:mleloss}
\end{align}
where $N_e$ is the size of the latent space representation vector and zero covariance is assumed in the $N_e$-variate normally distributed predictions.

Recent work by \cite{Sluijterman2023OptimalNetworks} has confirmed that mean-variance estimation neural networks benefit from a warm-up period in training where the variances are fixed and do not count toward the gradients within backpropagation, so we assumed $\forall_n\left(\sigma_{k+i}\right)_n=1$,
reducing the MLE loss to
\begin{align}
\mathcal{L}_{\text{MLE}}&=N_e\log\left(\sqrt{2\pi}\right)+\frac{1}{2N_f}\sum^{N_f}_{i=1}\sum_{n=1}^{N_e}\left[\left(z_{N_p+i}\right)_n-\left(\hat{z}_{N_p+i}\right)_n\right]^2,
\label{eq:mseloss}
\end{align}
which corresponds to optimizing the mean squared error. This simplified loss function is used during the first training iterations, either for a fixed amount of iterations (e.g., 1000) or until convergence. After warm-up, the training continues with the full MLE loss.

Note that the current combination of the encoder model, recurrent neural network model, and forecasting models forms an underdetermined optimization problem, where the optimal set of model parameters would lead to the encoder always producing the same latent space representations, regardless of the encoder input, and the recurrent neural network and forecasting models always predicting those representations. In the original CPC work, the contrastive loss encouraged the encoder to produce different representations for the positive and negative samples, solving this problem of underdetermination.

As MLE does not use negative samples, we added a decoder model $D$ to the pipeline with the sole task of reconstructing the original input $x_k$ from the latent space representation $z_k$, yielding $\hat{x}_k=D\left(z_k\right)$. This form of regularization encourages the encoder to produce meaningful representations. We chose the mean squared error as the additional loss term, yielding
\begin{align}
\mathcal{L}_{\text{MSE}}&=\frac{1}{N_p+N_f}\sum^{N_p+N_f}_{k=1}\text{MSE}\left(x_k,\hat{x}_k\right),
\end{align}
where $\text{MSE}$ is the mean squared error function that is appropriate for the data type and dimensionality of the input sample $x_k$. Together with the MLE loss, this gives the optimization problem
\begin{align}
\theta^*&=\operatornamewithlimits{argmin}_{\theta}\left(\mathcal{L}_{\text{MLE}}+\lambda\cdot\mathcal{L}_{\text{MSE}}\right),
\end{align}
where $\theta$ is the set of model parameters for models $E$, $G$, $F_1$, $F_2$, $\cdots$, $F_{N_f}$, and $D$, and $\lambda$ is the weight assigned to the mean squared error. After training, the decoder $D$ is no longer needed for the anomalous change point detection and can be discarded.

\subsection{Probability of conformance}
\label{section:probconf}
To determine whether the data element $x_{N_p+i}$ was expected given $x_1, x_2, \cdots, x_{N_p}$, we use the Mahalanobis distance $d_M$ between $\hat{z}_{N_p+i}$ and $z_{N_p+i}$,
\begin{align}
d_{z, N_p+i}&=d_M\left(z_{N_p+i}, \hat{z}_{N_p+i}, \sigma_{N_p+i}\right) \nonumber \\
&=\sqrt{\sum^{N_e}_{n=1}\frac{\left[\left(z_{N_p+i}\right)_n - \left(\hat{z}_{N_p+i}\right)_n\right]^2}{\left(\sigma_{N_p+i}\right)^2_n}}.
\end{align}
Whenever we consider this distance as a function of $Z_{N_p+i}$, the distance becomes a random variable $D_{z,N_p+i}=d_M\left(Z_{N_p+i}, \hat{z}_{N_p+i}, \sigma_{N_p+i}\right)$. Note that the squared Mahalanobis distance is a sum of squared independent univariate standard distributions, and therefore distributed according to a chi-squared distribution, i.e., $D^2_{z,N_p+i}\sim\chi^2\left(N_e\right)$.

Now, by using the corresponding cumulative distribution $F_{D^2_{z,N_p+i}}$, we can determine the probability of conformance in the latent space $p_{z, N_p+i}$ with
\begin{align}
p_{z,N_p+i}&=\textrm{P}\left(D^2_{z,N_p+i}>d^2_{z, N_p+i}\right)\nonumber\\
&= 1-F_{D^2_{z,N_p+i}}\left(d^2_{z, N_p+i},N_e\right).
\end{align}
Note that in the case of $N_e=1$, this reduces to 
\begin{align}
p_{z,N_p+i} &= 2\cdot Q\left(d^2_{z, N_p+i}\right),
\end{align}
where $Q\left(\cdot\right)$ is the tail distribution function for a univariate standard distribution. The pipeline functions as a binary classifier whenever thresholding is applied to the probability of conformance in the latent space, i.e., $z_{N_p+i}$ is anomalous if $p_{z,N_p+i}<\alpha$. In that case, the binary classifier rejects latent space representations with a probability of $\alpha$.

Now, if we consider the relationship with the probability of conformance in the data space, $p_{x,N_p+i}$, we must note that the relationship between $f\left(z_{N_p+i}|\hat{z}_{N_p+i},\sigma_{N_p+i}\right)$ and $p_{z,N_p+i}$ is strictly monotonically increasing. The same can trivially be said about the relationship between $f\left(x_{N_p+i}|x_1, x_2, \cdots, x_{N_p}\right)$ and $p_{x,N_p+i}$. Due to the proportionality between $f\left(z_{N_p+i}|\hat{z}_{N_p+i},\sigma_{N_p+i}\right)$ and $f\left(x_{N_p+i}|x_1, x_2, \cdots, x_{N_p}\right)$ (see section \ref{section:proxy}), we can therefore conclude that the relationship between $p_{z,N_p+i}$ and $p_{x,N_p+i}$ is also strictly monotonically increasing. This indicates that the ordering of values of $p_{x,N_p+i}$ is preserved in its proxy $p_{z,N_p+i}$, making this estimated probability of conformance in the latent space a useful metric for anomaly detection.

\section{Experiments}

\subsection{Proportionality test}
\label{section:propexp}
To determine whether the assumption of proportionality (see section \ref{section:proxy}), i.e., $f\left(x_{N_p+i}|x_1, x_2, \cdots, x_{N_p}\right)=K\cdot f\left(z_{N_p+i}|\hat{z}_{N_p+i},\sigma_{N_p+i}\right)$ with constant $K$, holds, we conducted a relatively simple experiment with a well-defined data distribution. In this experiment, $N_p=N_f=1$, $x_1\in\{-10, 0, 10\}$, $X_2\sim\mathcal{N}\left(\mu_{X_2}(x_1),\sigma^2_{X_2}(x_1)\right)$, $\mu_{X_2}(x_1)=x_1$, and $\sigma_{X_2}(x_1)=0.1\cdot x_1+2$. With this experiment, we challenged the pipeline to capture the data distributions for three different input data instances.

The pipeline consists of an encoder with a single fully connected layer to a latent space encoding with a size of $N_e=4$ and a decoder with a fully connected layer back to a single value. A gated recurrent unit (GRU) model with $N_g=8$ units is used for predictions together with a forecasting model consisting of six fully connected layers with $[16, 16, 32, 32, 64, 64]$ units, followed by two separate fully connected layers for the estimation of $\hat{z}_2$ (linear activation) and $\sigma_2$ (exponential activation). Training is performed using the RMSprop optimizer ($lr=1\cdot10^{-4}$, $\rho=0.9$) with a batch size of 64. The models are optimized for 1000 warm-up iterations based on the loss in Equation \ref{eq:mseloss}, after which training with the normal loss function (Equation \ref{eq:mleloss}) continues until convergence. The loss weight for the reconstruction loss is set to $\lambda=100$. Training is repeated 100 times to determine sensitivity to initial conditions, resulting in 100 models.

\begin{figure}[h]
\centering
\includegraphics[width=1.0\textwidth, trim=0cm 0.7cm 0cm 0.5cm]{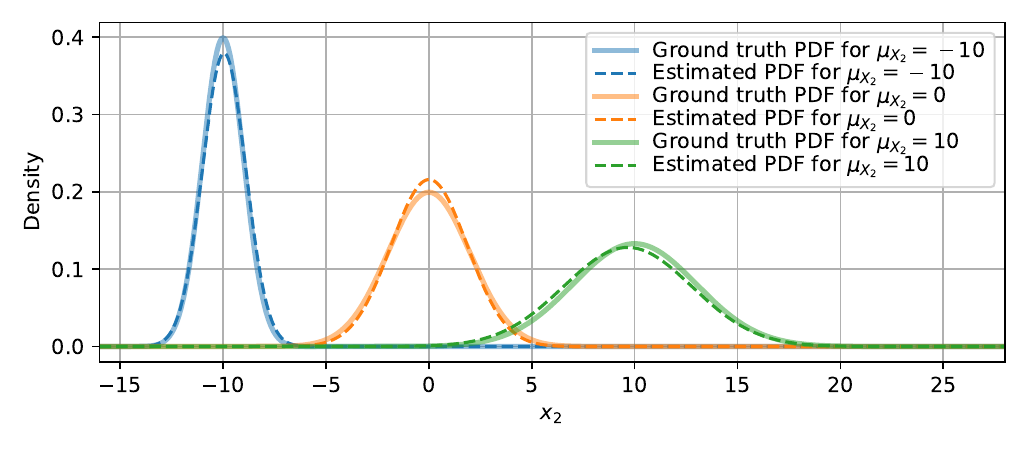}
\caption{An example of ground truth and estimated probability distribution functions for three different values of $x_1$.}\label{fig:proptest}
\end{figure}

After training, each model was evaluated for every combination of values $x_1\in\{-10, 0, 10\}$ and $x_2\in[-16, 28]$, where the range of $x_2$ is chosen to allow evaluation of values of $x_2$ within 6 standard deviations with respect to the ground truth data distribution means $\mu_{X_2}(x_1)=x_1$. The resulting probability functions were normalized to determine the probability density functions $K\cdot f\left(z_{N_p+i}|\hat{z}_{N_p+i},\sigma_{N_p+i}\right)$. Examples of the resulting functions are shown in Figure \ref{fig:proptest} with their corresponding ground truth data distributions. Considering that the shapes of Gaussian curves are sensitive to small deviations in parameters, we observe that the estimated PDFs tend to be similar to the ground truth data distributions.

To quantify the results, we fitted Gaussian functions to the obtained PDFs with zero residuals and compared the curve parameters $\hat{\mu}_{X_2}$ and $\hat{\sigma}_{X_2}$ to the ground truth data distribution parameters $\mu_{X_2}$ and $\sigma_{X_2}$ for the different values of $x_1$. The mean and standard deviation of these Gaussian curve parameters are shown in Table \ref{table:proptest}.

\begin{table}[h]
\caption{Quantitative results of the proportionality test, consisting of true and estimated PDF parameters for different values of $x_1$.}
\label{table:proptest}
\begin{tabular}{@{}rrrr@{}}
\toprule
$\mu_{X_2}=x_1$ & $\sigma_{X_2}$  & $\hat{\mu}_{X_2}$ ($\mu\pm\sigma$, $N=100$) & $\hat{\sigma}_{X_2}$ ($\mu\pm\sigma$, $N=100$)\\
\midrule
$-10$ & $1$ & $-10.04\pm 0.16$ & $1.07\pm 0.04$ \\
$0$   & $2$ & $-0.02\pm 0.13$ & $1.84\pm 0.06$ \\
$10$  & $3$ & $9.66\pm 0.19$ & $3.11\pm 0.10$ \\
\botrule
\end{tabular}
\end{table}

We can see that the ground truth and estimated parameters are almost identical in value, with relatively low standard deviations and thus low sensitivity to initial conditions. This indicates that the assumption of proportional probability density functions in the data and latent spaces holds.
 
\subsection{Sine wave frequency deviation}
\label{section:sineexp}
As a first practical demonstration of the pipeline's ability to detect anomalies, we synthesized a dataset of univariate signals in the form of (distorted) sine waves. The objective of the pipeline is to detect significant change points in sine wave frequency throughout the signal, where frequency-domain-based methods would not be able to.

To synthesize the appropriate training, validation and test datasets, we used a data-generating model $s(t)$ that introduces small variations in instantaneous frequency, amplitude, and baseline, as well as additive noise. For every signal, a center frequency before ($f_{c,before}$) and after the change point ($f_{c,after}$) is chosen within the interval $[0.5, 10]$ Hz, resulting in the instantaneous frequency
\begin{align}
    f_c(t) &=
    \begin{cases}
        f_{c,before} & \text{if } t \text{ before change point} \\
        f_{c,after} & \text{if } t \text{ after change point} 
    \end{cases}.
\end{align}

A signal containing the instantaneous frequencies at every time step $f(t)$ is generated using a constrained random walk, such that $f_c(t)-\frac{f_b}{2}\leq f(t)\leq f_c(t)+\frac{f_b}{2}$, where $f_b=0.25$ Hz is the frequency band. Furthermore, bounded random walks were used to create a baseline wander $-1\leq b(t) \leq 1$ and wave amplitude $0.5\leq a(t)\leq 2$. Lastly, additive noise was $n(t)\sim\mathcal{N}\left(0, \sigma_n^2\right)$, where $\sigma_n\sim\mathcal{U}(0, 0.2)$. The corresponding sine wave signal was then determined by
\begin{align}
    s(t) = a(t)\cdot\sin\left(2\pi\int_{0}^{t}f(t)dt\right) + b(t) + n(t).
\end{align}

\begin{figure}[h]
\centering
\includegraphics[width=1.0\textwidth, trim=0cm 1.0cm 0cm 0.3cm]{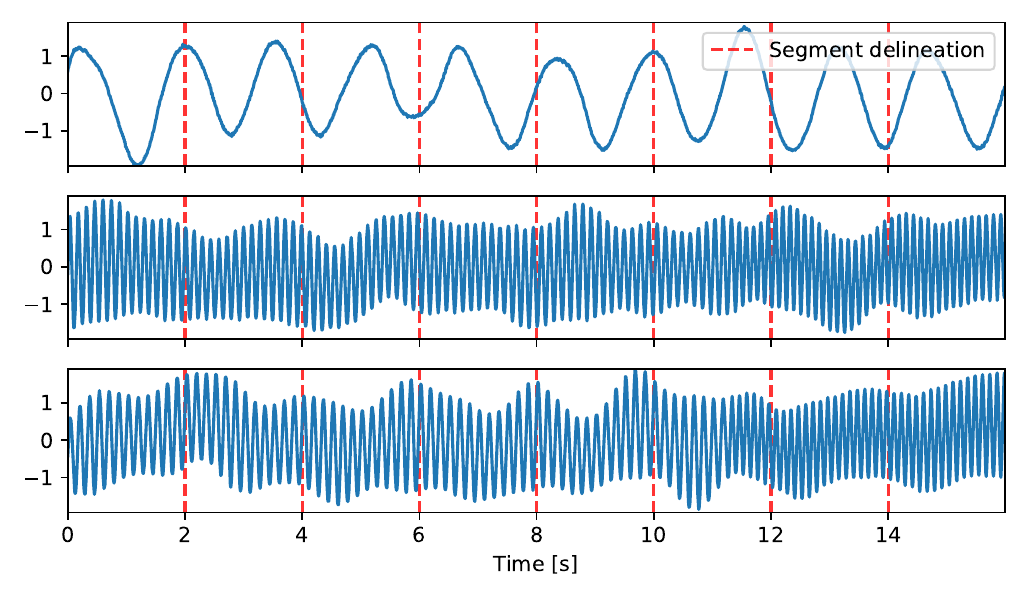}
\caption{Examples of generated signals. The top signal has no frequency deviation. The center signal has a frequency change of $1$ Hz at the change point around $t=11$ s. The bottom signal has a frequency change of $2$ Hz at the change point around $t=11$ s.}\label{fig:sine-examples}
\end{figure}

Using this data model, we synthesized a training and validation dataset of 1,000,000 and 100,000 signals respectively at a sampling frequency of $f_s=128$ Hz. The signals consist of 8 segments of 256 samples each, totaling 2048 samples (16 seconds). The first $5$ segments $x_1$ to $x_5$ are used to predict the latent space representations of the last $3$ segments $x_6$ to $x_8$. The center frequencies were set such that $f_{c,before}=f_{c,after}$, implying the absence of change points. In Figure \ref{fig:sine-examples}, examples of normal and abnormal signals are shown.

We used an encoder model $E$ with two blocks (convolutional, ReLU, BatchNorm, Maxpooling) with an increasing amount of filters (32 and 64), followed by a fully connected layer to a latent space with size $N_e=16$. The accompanying decoder $D$ follows a similar structure with a fully connected layer, followed by two blocks (convolutional, ReLU, upsampling) and one final convolutional layer. As the recurrent neural network $G$, we used a GRU with 32 units. The three forecasting models $F_1$, $F_2$, and $F_3$ consist of three fully connected layers (64, 128, and 256 units), followed by two separate fully connected layers for the estimation of $\hat{z}_6$ to $\hat{z}_8$ (linear activation) and $\sigma_6$ to $\sigma_8$ (exponential activation).

All models were simultaneously optimized on an Intel Core i9-10980XE CPU and NVIDIA RTX A6000 GPU hardware set-up with Windows Server 2022, Python 3.9.15 and TensorFlow 2.10. An RMSprop optimizer with a learning rate of $1\cdot10^{-4}$ and a batch size of $32$ was used. Training consisted of 1000 warm-up iterations ($0.89$ seconds/iteration), followed by normal training until convergence on the validation dataset ($\sim 365,000$ iterations, $0.58$ seconds/iteration). The reconstruction loss weight was set to $\lambda=10^4$. The model optimization required $\sim 1800$ MB of VRAM.

\begin{figure}[h]
\centering
\includegraphics[width=1.0\textwidth, trim=0cm 0.7cm 0cm 1.2cm]{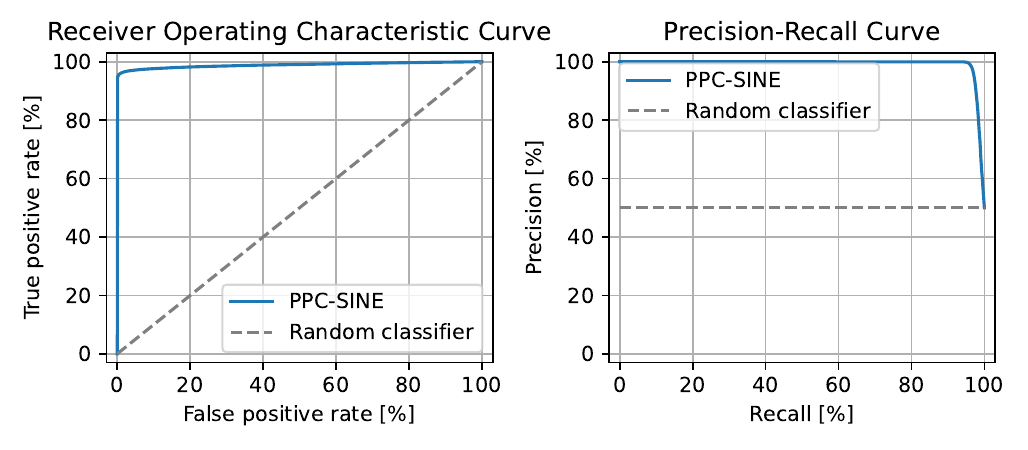}
\caption{Receiver Operating Characteristic and Precision-Recall curves for the application of the PPC pipeline on the sine wave frequency deviation example.}\label{fig:sine-roc-curve}
\end{figure}

After training, classification performance was tested on two test datasets. Both were generated in a similar manner as the training and validation set, except that change points could be introduced in segment $x_6$. Both datasets consisted of $100,000$ signals with anomalous change points and $100,000$ without anomalous change points. The first test dataset was used to determine an optimal classification threshold that maximizes the F1 score. The second test dataset, together with the determined classification threshold, was used to determine several classification performance metrics, which are provided in Table \ref{table:expresults}. Receiver Operating Characteristic (ROC) and Precision-Recall (PR) curves are shown in Figure \ref{fig:sine-roc-curve}. The classification performance metrics, as well as the ROC/PR curves, suggest that the PPC pipeline has a strong discriminative ability. Furthermore, inference time and memory were measured on the same machine, resulting in $0.098$ seconds/sample and $\sim 1200$ MB of VRAM.

\begin{table}[h]
\caption{Anomalous change point classification performance metrics for the \textit{Sine wave frequency deviation} experiment (section \ref{section:sineexp}) and \textit{Counting with MNIST digits} experiment (section \ref{section:mnistexp}).}
\label{table:expresults}
\begin{tabular}{@{}lrr@{}}
\toprule
Metric                           & \makecell{PPC-SINE \\ (Section \ref{section:sineexp})} & \makecell{PPC-MNIST \\ (Section \ref{section:mnistexp})}  \\
\midrule
ROC AUC                          & $0.989$  & $0.946$     \\
PR AUC                           & $0.992$  & $0.993$     \\
TP (True Positives)              & $95910$  & $88347875$  \\
FP (False Positives)             & $685$    & $3925349$   \\
TN (True Negatives)              & $99315$  & $6080573$   \\
FN (False Negatives)             & $4090$   & $1646203$   \\
Recall                           & $95.9\%$ & $98.2\%$    \\
Precision                        & $99.3\%$ & $95.7\%$    \\
Specificity                      & $99.3\%$ & $60.8\%$    \\
Balanced accuracy                & $97.6\%$ & $79.5\%$    \\
Matthews Correlation Coefficient & $0.953$  & $0.662$     \\
F1 Score                         & $0.976$  & $0.969$     \\
\botrule
\end{tabular}
\end{table}

\begin{figure}[h]
\centering
\includegraphics[width=1.0\textwidth, trim=0cm 0.7cm 0cm 0.3cm]{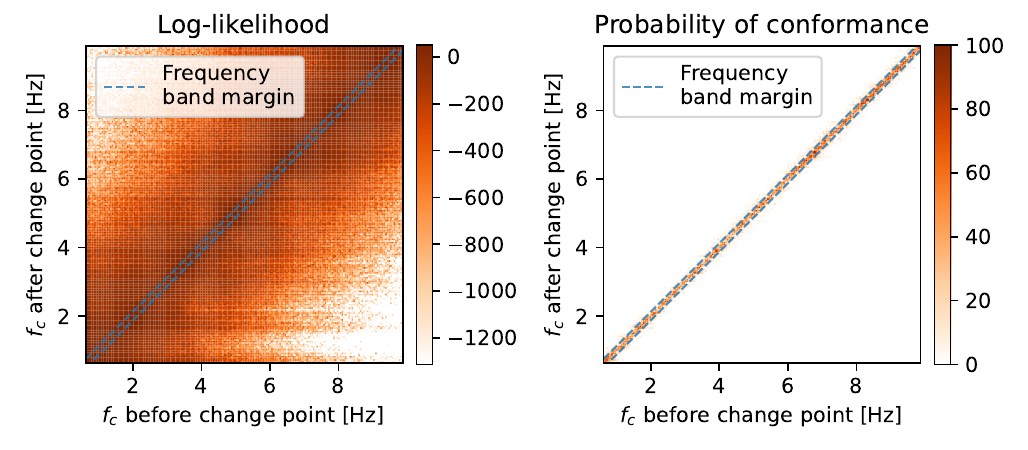}
\caption{Average log-likelihood (left) and probability of conformance (right) per combination of center frequencies. The frequency band margins indicate how much frequencies may fluctuate as part of the generated instantaneous frequencies $f(t)$.}\label{fig:sine-matrix}
\end{figure}

Furthermore, to visualize the response to different pairs of $f_{c,before}$ and $f_{c,after}$, we synthesized 10 signals for every combination of these center frequencies, with a resolution of $0.05$ Hz. This results in $\left(\frac{10-0.5}{0.05}\right)^2\cdot10=361,000$ signals. We applied the model to these signals and calculated the log-likelihood and probability of conformance for every pair of center frequencies $\left(f_{c,before},f_{c,after}\right)$. These metrics were averaged and are shown in Figure \ref{fig:sine-matrix}, together with the frequency band with which the training and validation signals were generated. As expected, the probability of conformance is high if $f_{c,before}\approx f_{c,after}$, and low for other combinations of center frequencies. This response suggests that the PPC pipeline is able to detect anomalous change points over a wide range of center frequencies.

\subsection{Counting with MNIST digits}
\label{section:mnistexp}
\begin{figure}[h]
\centering 
\includegraphics[width=1.0\textwidth, trim=0cm 0cm 0cm 0cm]{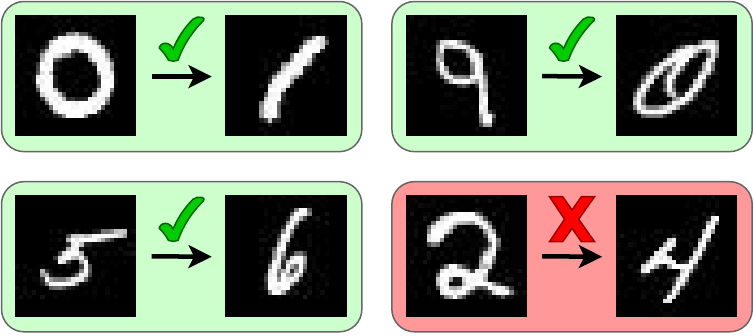}
\caption{Examples of correct (green) and incorrect (red) sequences of MNIST digits. During training, only correct sequences of digits are used.}\label{fig:mnist-digits}
\end{figure}

In this experiment, we used the Modified National Institute of Standards and Technology (MNIST) dataset \citep{Lecun1998Gradient-BasedRecognition}, consisting of 60,000 training images and 10,000 test images of handwritten digits. The objective within this experiment for the pipeline is to determine whether the corresponding digits of a set of two MNIST images are sequential, e.g., if $x_1$ formed an image of the digit 3, $x_2$ should represent the digit 4. In the special case of digit 9, we considered 0 to be its successor. Examples of sequences that (do not) follow this pattern are provided in Figure \ref{fig:mnist-digits}. The experiment was designed to determine if the pipeline is able to capture (deviations from) patterns in relatively complex sequential image data.

First, 10,000 validation images were selected from the original set of training images, reflecting the class composition of the training set. All image pixels were normalized to the range $[0, 1]$. Data augmentation is applied online, consisting of random rotation (between $-18^{\circ}$ and $18^{\circ}$), random rectangular zooming (between $-5\%$ and $5\%$), and random translation (between $-5\%$ and $5\%$).

We used an encoder model $E$ with two blocks (convolutional, ReLU, BatchNorm, Maxpooling) with an increasing amount of filters (32 and 64), followed by a fully connected layer to a latent space with size $N_e=16$. The accompanying decoder $D$ follows a similar structure with a fully connected layer, followed by two blocks (convolutional, ReLU, upsampling) and one final convolutional layer. As the recurrent neural network $G$, we used a GRU with 32 units. The single forecasting model $F_1$ consists of three fully connected layers (64, 128, and 256 units), followed by two separate fully connected layers for the estimation of $\hat{z}_2$ (linear activation) and $\sigma_2$ (exponential activation).

All models were simultaneously optimized on an Intel Core i9-10980XE CPU and NVIDIA RTX A6000 GPU hardware set-up with Windows Server 2022, Python 3.9.15 and TensorFlow 2.10. An RMSprop optimizer with a learning rate of $1\cdot10^{-4}$ and a batch size of $32$ was used. Training consisted of 1000 warm-up iterations ($0.57$ seconds/iteration), followed by normal training until convergence ($\sim 132,000$ iterations, $0.57$ seconds/iteration. The reconstruction loss weight was set to $\lambda=10^4$. The model optimization required $\sim 1300$ MB of VRAM.

\begin{figure}[h]
\centering
\includegraphics[width=1.0\textwidth, trim=0cm 0.7cm 0cm 1.2cm]{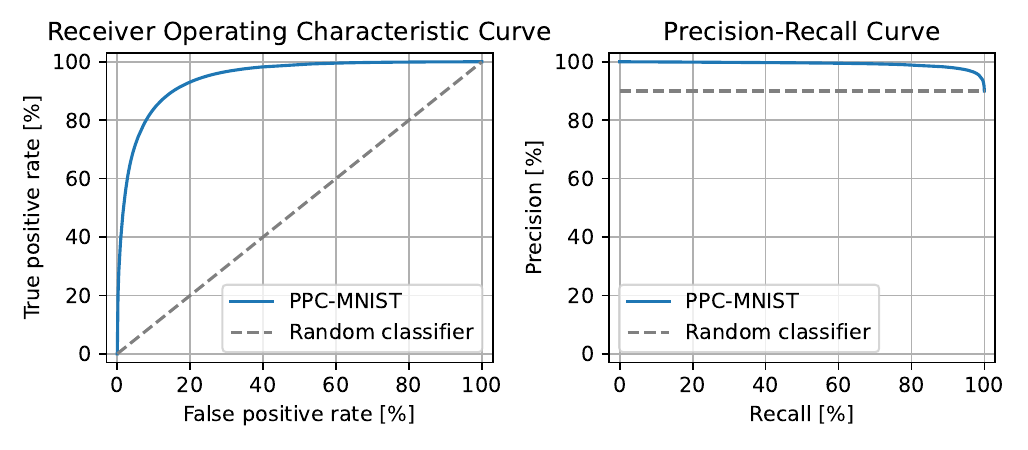}
\caption{Receiver Operating Characteristic and Precision-Recall curves for the application of the PPC pipeline on the successive MNIST digits example.}\label{fig:mnist-roc-curve}
\end{figure}

After training, the pipeline was used to compare every pair of two images in the test set, yielding a log-likelihood and probability of conformance for every pair. Similarly to the \textit{Sine wave frequency deviation} experiment (Section \ref{section:sineexp}), a classification threshold was determined on the validation set, and classification performance metrics were calculated based on the test set. These metrics are reported in Table \ref{table:expresults} and suggest strong discriminative ability of the PPC pipeline. ROC and PR curves are shown in Figure \ref{fig:mnist-roc-curve}. Inference took $0.026$ seconds/sample and $\sim 1100$ MB of VRAM.

\begin{figure}[h]
\centering 
\includegraphics[width=1.0\textwidth, trim=0cm 0.7cm 0cm 0.3cm]{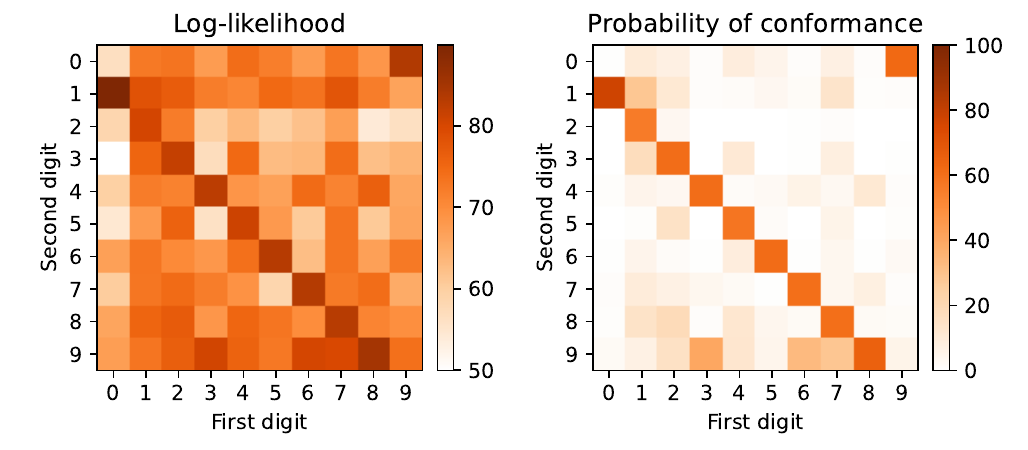}
\caption{Average log-likelihood (left) and probability of conformance (right) per combination of digits.}\label{fig:mnist-matrix}
\end{figure}

To see the pipeline's response to different combinations of digits, log-likelihood and probability of conformance were averaged over every pair of digits and are shown in Figure \ref{fig:mnist-matrix}. As expected, the log-likelihoods and probabilities of conformance on the subdiagonal and top right are significantly higher than for other combinations of digits. This indicates that the model indeed is able to differentiate between successive digits and non-successive digits. Some combinations, like the combination for $(3,9)$, show higher values due to the similarity between digits, e.g., a handwritten 4 is similar in shape to the digit 9.

Furthermore, to demonstrate the ability of the pipeline to learn latent space distributions, we applied principal component analysis (PCA) dimensionality reduction on the encodings of the test images to create 2D representations. We inferred the predicted distributions for each digit, reduced these distribution parameters to bivariate Gaussian distribution parameters using the same PCA model, and averaged the parameters over all inferred distributions.

\begin{figure}[h]
\centering
\includegraphics[width=1.0\textwidth, trim=0cm 1.2cm 0cm 0.8cm]{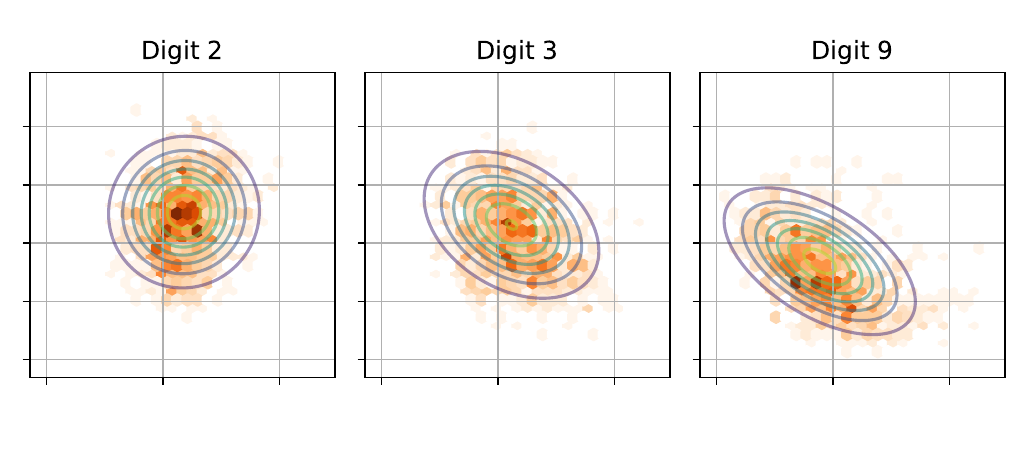}
\caption{Hexagonal binning plots for the PCA-reduced latent space representations of different test images and digits, with contours of the inferred bivariate distributions projected over these plots.}\label{fig:mnist-latent-hexbin}
\end{figure}

In Figure \ref{fig:mnist-latent-hexbin}, we show hexagonal binning plots for the reduced latent space representations for three example digits, and project the contours of the corresponding inferred bivariate distributions over these latent spaces. While PCA dimensionality reduction tends to remove essential information in the latent space, it can still be observed that the general shape of the distribution fits the shape of the hexagonal binning plots. This indicates that the pipeline is able to align the latent space and the predicted distributions, which is essential for anomalous change point detection.

\subsection{Artifact detection for MRSI}
\label{section:mrsiexp}
In this last experiment, we demonstrate the applicability of the proposed method to in-vivo proton magnetic resonance spectroscopic imaging (MRSI) data. MRSI allows non-invasive measurement and visualization of the chemical composition of tissues and has become a valuable tool for diagnosis and treatment monitoring of various diseases, such as cancer, neurological disorders, and traumatic brain injuries \citep{Posse2013MRAdvances}. However, clinical adoption of MRSI remains limited due to long acquisition times, low spatial resolutions, and variable spectral quality \citep{Maudsley2021AdvancedRecommendations}. Furthermore, the detection of artifacts is an active and challenging area of research \citep{Kreis2004IssuesArtifacts}, specifically the automation thereof \citep{PedrosadeBarros2017ImprovingData, Gurbani2018AMRI, Kyathanahally2018QualityTools, Jang2021UnsupervisedBrain, vandeSande2023AWorkflow}.

While metabolite signals and their characteristics are well understood and commonly modeled using density matrix simulations \citep{Zhang2017FastSpectroscopy}, other signal contributions, like macromolecules, water/fat residuals, and other nuisance signals \citep{Kreis2004IssuesArtifacts} are anomalous due to sub-optimal localization performance and hardware imperfections \citep{Kreis2021TerminologyRecommendations}. Therefore, the proposed PPC pipeline was trained and validated with predictable synthetic single voxel spectroscopy (SVS) datasets and tested on in-vivo spectra of MRSI measurements. The simulated spectra were obtained using density matrix simulated metabolite signals with concentration ranges as reported for normal adult brains \citep{DeGraaf2019InTechniques}. To incorporate field inhomogeneities into the simulations, Lorentzian and Gaussian broadening were applied. In-vivo spectra were extracted from the MRSI database reported in \cite{Bhogal2020LipidsuppressedT} as part of a study at the University Medical Center Utrecht. The data were acquired using a free-induction decay (FID) sequence at 7 Tesla with $5 \times 5 \times 10$ mm sized voxels. Full data acquisition details are provided in the methods section of \cite{Bhogal2020LipidsuppressedT}.

The model takes a segment $x_1$ (32 samples) of the SVS signal in the frequency domain and predicts the latent space representations of the next segment $x_2$. We used an encoder model $E$ with four fully connected layers (512 units and ReLU activation), followed by a final linear layer to a latent space with size $N_e=16$. The accompanying decoder $D$ follows an identical structure with the final linear layer mapping back to the segment width. The recurrent neural network $G$ consists of 32 units and the forecasting models $F_1$ has 6 fully connected layers (64, 64, 128, 128, 256, and 256 units), followed by two separate fully connected layers for the estimation of $\hat{z}_2$ (linear activation) and $\sigma_2$ (exponential activation). All models were optimized simultaneously using the RMSprop optimizer with a learning rate of $1\cdot10^{-4}$ and a batch size of 256. Training consisted of $2\cdot10^4$ warm-up iterations, followed by normal training until convergence ($2\cdot10^6$ iterations). The reconstruction loss weight was set to $\lambda=10^5$.

\begin{figure}[h]
\centering
\includegraphics[width=1.0\textwidth, trim=0cm 1.0cm 0cm 0.8cm]{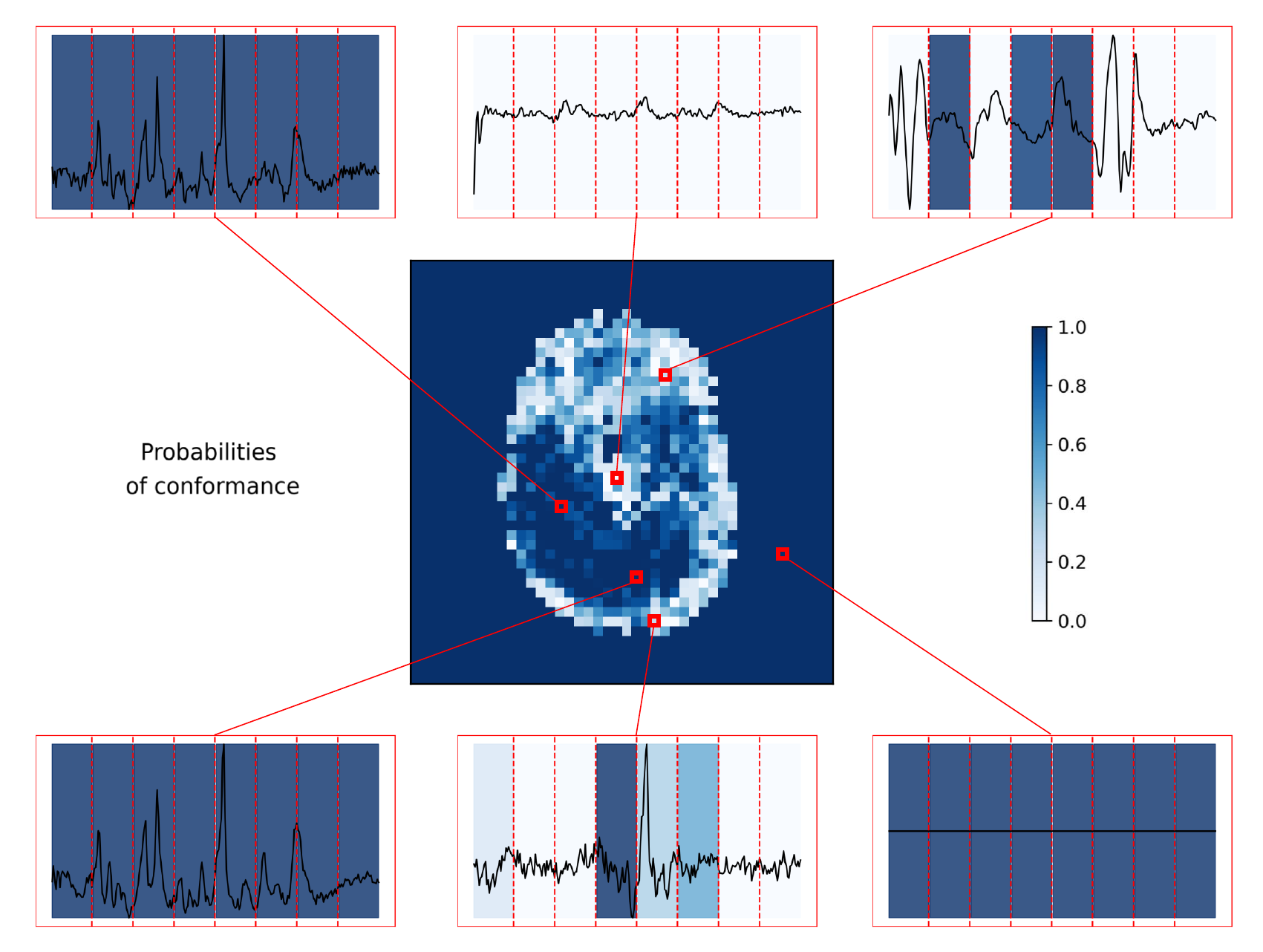}
\caption{In-vivo 2D MRSI slice with indicated locations of selected segmented example spectra. The probability of conformance is shown for each segment of the spectra as well as averaged for each pixel (representing a single voxel).}\label{fig:mrsi-prob}
\end{figure}

After training, the pipeline was used to predict all segments of every spectrum of an example in-vivo 2D MRSI slice. We further calculate the likelihoods and probabilities of conformance for every segment. Figure \ref{fig:mrsi-prob} depicts the probabilities of conformance for six example spectra located at the indicated locations as well as a probability map with pixels representing averaged spectra.
Moving from the top left to top right spectrum: spectra within the homogeneous regions of the brain appear normal and show the highest probabilities of conformance, measurements near the ventricles are distorted completely, and susceptibility artifacts coming from the nasal cavity are present and detected by the method. From bottom left to right: non-anomalous spectra are consistently labeled with high probabilities, spectra close to the skull show residual lipid signal artifacts and lower probabilities of conformance, and the spectra outside the brain mask are nulled leading to no anomalies.

\begin{figure}[h]
\centering
\includegraphics[width=1.0\textwidth, trim=0cm 1.0cm 0cm 0.8cm]{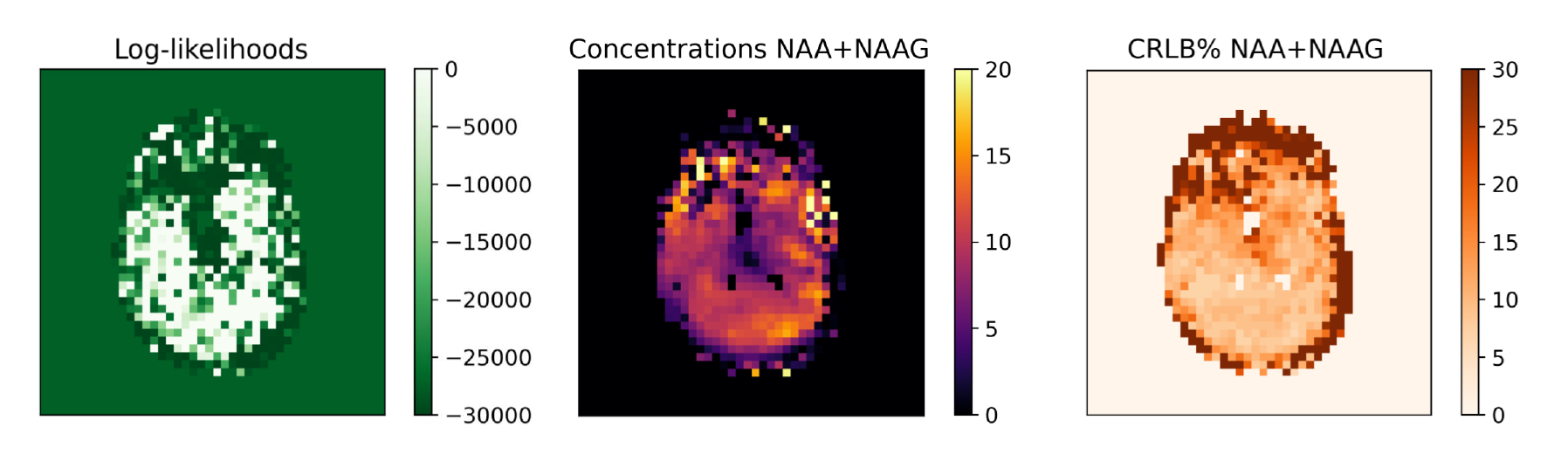}
\caption{2D brain maps for the estimated likelihoods, the metabolite concentration map for total N-acetylaspartate (NAA + NAAG) estimated by LCModel~\citep{Provencher1993EstimationSpectra}, as well as the corresponding CRLB\% (left to right).} \label{fig:mrsi-comp}
\end{figure}

Figure \ref{fig:mrsi-comp} shows the estimated log-likelihoods next to the metabolite concentration map for total N-acetylaspartate (NAA + NAAG), one of the most prominent spectral components \citep{Maudsley2021AdvancedRecommendations}, as well as next to the obtained Cram$\acute{\text{e}}$r-Rao lower bound percentage values (CRLB\%) for the metabolite concentration as estimated by the common quantification software LCModel \citep{Provencher1993EstimationSpectra}. The concentration estimates are obtained via least squares fitting of the measurements and a linear combination model of metabolite signals \citep{Near2021PreprocessingRecommendations}. The CRLB\% is calculated via the Fisher information matrix and represents a lower bound on the standard deviation of the estimates \citep{Landheer2021AreSpectroscopy}. It is commonly used as a quality measure for both SVS and MRSI \citep{Oz2014ClinicalDisorders, Kreis2021TerminologyRecommendations}. Comparing the CRLB\% with the estimated log-likelihoods and probabilities of conformance there is a clear inverse correlation, specifically concerning artifacts arising from the skull, the ventricles, and the nasal cavity.

\section{Discussion} \label{section:discussion}
In this work, we proposed a novel pipeline for anomalous change point detection that leverages the power of probabilistic predictions and latent space encoding to capture the underlying data distribution of non-anomalous data. Through several experiments, we demonstrated the effectiveness and wide applicability of our pipeline in detecting anomalous change points.

In contrast to some other methods mentioned in section \ref{section:background}, anomaly detection with the proposed PPC method has a linear time complexity ($\mathcal{O}(n)$) with respect to the number of samples $n$, as individual samples are processed independently. This allows for scalability to large datasets. While training the PPC pipeline can be computationally expensive compared to non-neural network approaches, the models used in our experiments are remarkably lightweight compared to other deep neural network architectures. The models used in sections \ref{section:sineexp}, \ref{section:mnistexp}, and \ref{section:mrsiexp} have $5.8\cdot10^5$, $2.3\cdot10^5$ and $9.5\cdot10^5$ trainable parameters, respectively, while modern deep neural networks have significantly more parameters. For example, Inceptionv3 \citep{Szegedy2016RethinkingVision}, ResNet-50 \citep{He2016DeepRecognition}, and VGG-16 \citep{Simonyan2014VeryRecognition} have about $2.4\cdot10^7$, $2.5\cdot10^7$, and $1.4\cdot10^8$ trainable parameters, respectively. Furthermore, where other methods may fail to provide meaningful or intuitive anomaly scores, PPC is able to provide an anomaly score that is not only normalized to a closed unit interval $[0, 1]$, but also indicates a probability of data conforming to what was expected.

With our first experiment (section \ref{section:propexp}), we have shown that the architectural prior within this pipeline, combined with MLE loss optimization, leads to an estimation of the latent space representation likelihoods that is proportional to the underlying data distribution. This forms a solid basis for the statistical meaning of the probability of conformance. The other experiments have shown that the pipeline can be applied to several synthetic and real-world problems with different data modalities. The \textit{sine wave frequency deviation} experiment (section \ref{section:sineexp}) shows excellent performance for anomaly detection in a univariate time series problem. The \textit{counting with MNIST digits} example (section \ref{section:mnistexp}) showed how the pipeline can discover sequential patterns in image data.

The \textit{artifact detection for MRSI} experiment (section \ref{section:mrsiexp}) shows the real-world application of the method to MRSI data. The probability of conformance delivers a valuable measure for the quality of MRSI spectra and enables the filtering of artifact-ridden spectra by setting a specific threshold. This is particularly useful with high‐resolution whole‐brain acquisition techniques where hundreds of thousands of spectra are acquired while only good-quality spectra can be quantified reliably \citep{Maudsley2006ComprehensiveImaging, Nam2023DeuteriumT}. Note the CRLB\% is only obtained after computationally intensive least-squares fitting of the spectra. In addition, having a segment-wise quality measure for each spectrum can allow metabolite-targeted filtering to improve studies of specific brain pathologies \citep{Bogner2021AcceleratedTechniques, Govindaraju2004VolumetricInjury., Pelletier20023-DBrain}.

Despite the promising results, our proposed method and the experiments have some limitations. The application of the pipeline on a specific problem requires the choice of a suitable encoder-decoder pair, not only to handle the specific data type but also to handle the context of the problem. For example, in the case of the sine wave example, it would be obvious to choose a segment length such that the receptive field of the encoder covers the largest wave period (2 seconds). Furthermore, a suitable latent space encoding dimension $N_e$ needs to be chosen, as to prevent both under- and overfitting. While choosing a suitable encoder/decoder may not be trivial, as a rule of thumb, we suggest adopting architectures that would yield reasonably low reconstruction losses in an autoencoder application. This would guarantee the inference of information-rich latent space representations that can be used for predictive modeling.

The dependency on the decoder to regularize the model and prevent the pipeline from becoming underdetermined is necessary but inconvenient. As the predictive modeling competes with the reconstruction modeling in a joint loss landscape, the latent space representations are not fully optimized for the predictive modeling, inevitably leading to diminished performance on anomaly detection. For future research, we suggest exploring the use of contrastive learning in either a semi-supervised setting (positive and negative samples both from non-anomalous data) or a fully supervised setting (positive samples from non-anomalous data, negative samples from normal and abnormal data). We expect that the use of contrastive learning conditions the latent space such that regularization with a decoder is no longer necessary, as contrastive learning incentivizes distinct latent space representations by design.

Lastly, we recognize the lack of benchmark evaluations in this work. The lack of available real-world datasets forms an obstacle to the development and testing of CPD and AD methods \citep{Pang2022DeepDetection}. This is even more problematic for ACPD methods such as PPC. All works mentioned in section \ref{section:ACPDmethods} use specifically-designed synthetic datasets, private datasets, datasets containing only normal data and anomalies (for testing AD methods), or datasets containing only normal data and abnormal change points (for CPD testing methods). Furthermore, these datasets are often low-complexity, univariate time series data that are outdated and/or unannotated. Therefore, we strongly recommend that future research on ACPD methods includes the development of publicly available benchmark datasets that form the standard for the evaluation of ACPD methods.

\section{Conclusion}
This study introduces a novel pipeline for anomalous change point detection that leverages an architectural prior with probabilistic predictions and latent space encoding, and a training method that captures the underlying data distributions. These distributions can then be used to determine an anomaly score in the form of a \textit{probability of conformance}.

Theoretically, this pipeline allows for wide application to different problems and data types. Where classical methods are generally not capable of dealing with highly complex data, and deep anomaly detection methods generally lack interpretability, meaningful anomaly scores, or low computational complexity, the proposed method solves these problems.

Through a series of experiments, we evaluated the effectiveness and versatility of our proposed method across various data modalities, including synthetic time series, image data, and real-world magnetic resonance spectroscopic imaging (MRSI) data. Our findings suggest that the pipeline is capable of accurately detecting anomalous change points in different problems with different types of data, showcasing its potential for a wide range of applications in anomaly detection and monitoring.

Despite the promising results, it is important to acknowledge the limitations of our approach, including the assumptions of data normality and the dependence on specific encoder-decoder architectures. Additionally, the lack of publicly available benchmark datasets poses a challenge for comprehensive evaluation and comparison with existing methods. Future research directions include exploring contrastive learning to alleviate the dependency on decoder regularization.

In summary, our study contributes to the growing body of research in anomaly detection by presenting a widely applicable pipeline that demonstrates promising performance across diverse data domains.

\section{Statements and declarations}

\noindent\textbf{Author contributions}: All authors contributed to the study conception and design. Theoretical development was performed by Roelof G. Hup. Experimental design and execution were performed by Roelof G. Hup and Julian P. Merkofer. Data acquisition and analysis were performed by Roelof G. Hup, Julian P. Merkofer, and Alex A. Bhogal. Funding was acquired by Rik Vullings, Reinder Haakma, and Ruud J.G. van Sloun. The first draft of the manuscript was written by Roelof G. Hup and Julian P. Merkofer and all authors commented on previous versions of the manuscript. All authors read and approved the final manuscript.
\\\\
\noindent\textbf{Funding}: This work is financed by the PPP Allowance made available by Top Sector Life Sciences \& Health to the Dutch Heart Foundation to stimulate public-private partnerships, grant number 01-003-2021-B005, Philips Electronics Nederland B.V., and Spectralligence (EUREKA IA Call, ITEA4 project 20209).
\\\\
\noindent\textbf{Competing interests}: Reinder Haakma is employed at Philips Electronics Nederland B.V. Rik Vullings has a consultancy position at Philips Electronics Nederland B.V.
\\\\
\noindent\textbf{Ethical approval}: In-vivo spectra were extracted from the database reported in \cite{Bhogal2020LipidsuppressedT} as part of a study approved by the medical research ethics committee of the University Medical Center Utrecht.
Experiments were conducted according to the regulations of the Dutch \textit{Medical Research Involving Human Subjects Act} and in concordance with the Declaration of Helsinki. The International Conference on Harmonization Good Clinical Practice (ICH-GCP) was used as guidance throughout all study-related activities. All participants provided written informed consent.
\\\\
\noindent\textbf{Data availability}: The synthetic data used in sections \ref{section:propexp}, \ref{section:sineexp}, and \ref{section:mrsiexp} can be generated on demand or are available upon request. The MNIST dataset in section \ref{section:mnistexp} is publicly available. The in-vivo MRSI data in section \ref{section:mrsiexp} can be made available upon submission of a project outline to co-author Alex A. Bhogal.

\bibliography{references}

\begin{thebibliography}{}
\renewcommand{\doi}[1]{\url{https://doi.org/#1}}
\bibcommenthead

\bibitem [\protect \citeauthoryear {%
Adams%
\ \BBA {} MacKay%
}{%
Adams%
\ \BBA {} MacKay%
}{%
{\protect \APACyear {2007}}%
}]{%
Adams2007BayesianDetection}
\APACinsertmetastar {%
Adams2007BayesianDetection}%
\begin{APACrefauthors}%
Adams, R.P.%
\BCBT {}\ \BBA {} MacKay, D.J.C.%
\end{APACrefauthors}%
\unskip\
\newblock
\APACrefYearMonthDay{2007}{10}{}.
\newblock
{\BBOQ}\APACrefatitle {{Bayesian Online Changepoint Detection}} {{Bayesian Online Changepoint Detection}}.{\BBCQ}
\newblock
\APACjournalVolNumPages{arXiv}{}{}{,}
\newblock

\newblock

\PrintBackRefs{\CurrentBib}

\bibitem [\protect \citeauthoryear {%
Apostol%
, Truic{\u{a}}%
, Pop%
\BCBL {}\ \BBA {} Esposito%
}{%
Apostol%
\ \protect \BOthers {.}}{%
{\protect \APACyear {2021}}%
}]{%
Apostol2021ChangeData}
\APACinsertmetastar {%
Apostol2021ChangeData}%
\begin{APACrefauthors}%
Apostol, E\BHBI S.%
, Truic{\u{a}}, C\BHBI O.%
, Pop, F.%
\BCBL {} Esposito, C.%
\end{APACrefauthors}%
\unskip\
\newblock
\APACrefYearMonthDay{2021}{6}{}.
\newblock
{\BBOQ}\APACrefatitle {{Change Point Enhanced Anomaly Detection for IoT Time Series Data}} {{Change Point Enhanced Anomaly Detection for IoT Time Series Data}}.{\BBCQ}
\newblock
\APACjournalVolNumPages{Water}{13}{12}{1633,}
\newblock
\begin{APACrefDOI} \doi{10.3390/w13121633} \end{APACrefDOI}
\newblock

\newblock

\PrintBackRefs{\CurrentBib}

\bibitem [\protect \citeauthoryear {%
Atashgahi%
, Mocanu%
, Veldhuis%
\BCBL {}\ \BBA {} Pechenizkiy%
}{%
Atashgahi%
\ \protect \BOthers {.}}{%
{\protect \APACyear {2023}}%
}]{%
Atashgahi2023Memory-freeApproach}
\APACinsertmetastar {%
Atashgahi2023Memory-freeApproach}%
\begin{APACrefauthors}%
Atashgahi, Z.%
, Mocanu, D.C.%
, Veldhuis, R.%
\BCBL {} Pechenizkiy, M.%
\end{APACrefauthors}%
\unskip\
\newblock
\APACrefYearMonthDay{2023}{12}{}.
\newblock
{\BBOQ}\APACrefatitle {{Memory-free Online Change-point Detection: A Novel Neural Network Approach}} {{Memory-free Online Change-point Detection: A Novel Neural Network Approach}}.{\BBCQ}
\newblock
\APACjournalVolNumPages{arXiv}{}{}{,}
\newblock

\newblock

\PrintBackRefs{\CurrentBib}

\bibitem [\protect \citeauthoryear {%
Audibert%
, Michiardi%
, Guyard%
, Marti%
\BCBL {}\ \BBA {} Zuluaga%
}{%
Audibert%
\ \protect \BOthers {.}}{%
{\protect \APACyear {2020}}%
}]{%
Audibert2020USAD:Series}
\APACinsertmetastar {%
Audibert2020USAD:Series}%
\begin{APACrefauthors}%
Audibert, J.%
, Michiardi, P.%
, Guyard, F.%
, Marti, S.%
\BCBL {} Zuluaga, M.A.%
\end{APACrefauthors}%
\unskip\
\newblock
\APACrefYearMonthDay{2020}{8}{}.
\newblock
{\BBOQ}\APACrefatitle {{USAD: UnSupervised Anomaly Detection on Multivariate Time Series}} {{USAD: UnSupervised Anomaly Detection on Multivariate Time Series}}.{\BBCQ}
\newblock
 \APACrefbtitle {Proceedings of the ACM SIGKDD International Conference on Knowledge Discovery and Data Mining} {Proceedings of the acm sigkdd international conference on knowledge discovery and data mining}\ (\BPGS\ 3395--3404).
\newblock
\APACaddressPublisher{}{Association for Computing Machinery}.
\PrintBackRefs{\CurrentBib}

\bibitem [\protect \citeauthoryear {%
Bai%
\ \BBA {} Perron%
}{%
Bai%
\ \BBA {} Perron%
}{%
{\protect \APACyear {1998}}%
}]{%
Bai1998EstimatingChanges}
\APACinsertmetastar {%
Bai1998EstimatingChanges}%
\begin{APACrefauthors}%
Bai, J.%
\BCBT {}\ \BBA {} Perron, P.%
\end{APACrefauthors}%
\unskip\
\newblock
\APACrefYearMonthDay{1998}{1}{}.
\newblock
{\BBOQ}\APACrefatitle {{Estimating and Testing Linear Models with Multiple Structural Changes}} {{Estimating and Testing Linear Models with Multiple Structural Changes}}.{\BBCQ}
\newblock
\APACjournalVolNumPages{Econometrica}{66}{1}{47,}
\newblock
\begin{APACrefDOI} \doi{10.2307/2998540} \end{APACrefDOI}
\newblock

\newblock

\PrintBackRefs{\CurrentBib}

\bibitem [\protect \citeauthoryear {%
Bashar%
\ \BBA {} Nayak%
}{%
Bashar%
\ \BBA {} Nayak%
}{%
{\protect \APACyear {2020}}%
}]{%
Bashar2020TAnoGAN:Networks}
\APACinsertmetastar {%
Bashar2020TAnoGAN:Networks}%
\begin{APACrefauthors}%
Bashar, M.A.%
\BCBT {}\ \BBA {} Nayak, R.%
\end{APACrefauthors}%
\unskip\
\newblock
\APACrefYearMonthDay{2020}{12}{}.
\newblock
{\BBOQ}\APACrefatitle {{TAnoGAN: Time Series Anomaly Detection with Generative Adversarial Networks}} {{TAnoGAN: Time Series Anomaly Detection with Generative Adversarial Networks}}.{\BBCQ}
\newblock
 \APACrefbtitle {2020 IEEE Symposium Series on Computational Intelligence (SSCI)} {2020 ieee symposium series on computational intelligence (ssci)}\ (\BPGS\ 1778--1785).
\newblock
\APACaddressPublisher{}{IEEE}.
\PrintBackRefs{\CurrentBib}

\bibitem [\protect \citeauthoryear {%
Bhogal%
\ \protect \BOthers {.}}{%
Bhogal%
\ \protect \BOthers {.}}{%
{\protect \APACyear {2020}}%
}]{%
Bhogal2020LipidsuppressedT}
\APACinsertmetastar {%
Bhogal2020LipidsuppressedT}%
\begin{APACrefauthors}%
Bhogal, A.A.%
, Broeders, T.A.A.%
, Morsinkhof, L.%
, Edens, M.%
, Nassirpour, S.%
, Chang, P.%
\BDBL {}Wijnen, J.P.%
\end{APACrefauthors}%
\unskip\
\newblock
\APACrefYearMonthDay{2020}{12}{}.
\newblock
{\BBOQ}\APACrefatitle {{Lipid‐suppressed and tissue‐fraction corrected metabolic distributions in human central brain structures using 2D 1H magnetic resonance spectroscopic imaging at 7 T}} {{Lipid‐suppressed and tissue‐fraction corrected metabolic distributions in human central brain structures using 2D 1H magnetic resonance spectroscopic imaging at 7 T}}.{\BBCQ}
\newblock
\APACjournalVolNumPages{Brain and Behavior}{10}{12}{,}
\newblock
\begin{APACrefDOI} \doi{10.1002/brb3.1852} \end{APACrefDOI}
\newblock

\newblock

\PrintBackRefs{\CurrentBib}

\bibitem [\protect \citeauthoryear {%
Bleakley%
\ \BBA {} Vert%
}{%
Bleakley%
\ \BBA {} Vert%
}{%
{\protect \APACyear {2011}}%
}]{%
Bleakley2011TheDetection}
\APACinsertmetastar {%
Bleakley2011TheDetection}%
\begin{APACrefauthors}%
Bleakley, K.%
\BCBT {}\ \BBA {} Vert, J\BHBI P.%
\end{APACrefauthors}%
\unskip\
\newblock
\APACrefYearMonthDay{2011}{6}{}.
\newblock
{\BBOQ}\APACrefatitle {{The group fused Lasso for multiple change-point detection}} {{The group fused Lasso for multiple change-point detection}}.{\BBCQ}
\newblock
\APACjournalVolNumPages{arXiv}{}{}{,}
\newblock

\newblock

\PrintBackRefs{\CurrentBib}

\bibitem [\protect \citeauthoryear {%
Bogner%
, Otazo%
\BCBL {}\ \BBA {} Henning%
}{%
Bogner%
\ \protect \BOthers {.}}{%
{\protect \APACyear {2021}}%
}]{%
Bogner2021AcceleratedTechniques}
\APACinsertmetastar {%
Bogner2021AcceleratedTechniques}%
\begin{APACrefauthors}%
Bogner, W.%
, Otazo, R.%
\BCBL {} Henning, A.%
\end{APACrefauthors}%
\unskip\
\newblock
\APACrefYearMonthDay{2021}{5}{}.
\newblock
{\BBOQ}\APACrefatitle {{Accelerated MR spectroscopic imaging—a review of current and emerging techniques}} {{Accelerated MR spectroscopic imaging—a review of current and emerging techniques}}.{\BBCQ}
\newblock
\APACjournalVolNumPages{NMR in Biomedicine}{34}{5}{,}
\newblock
\begin{APACrefDOI} \doi{10.1002/nbm.4314} \end{APACrefDOI}
\newblock

\newblock

\PrintBackRefs{\CurrentBib}

\bibitem [\protect \citeauthoryear {%
Box%
, Jenkins%
, Reinsel%
\BCBL {}\ \BBA {} Ljung%
}{%
Box%
\ \protect \BOthers {.}}{%
{\protect \APACyear {2016}}%
}]{%
Box2016TimeControl}
\APACinsertmetastar {%
Box2016TimeControl}%
\begin{APACrefauthors}%
Box, G.E.P.%
, Jenkins, G.M.%
, Reinsel, G.C.%
\BCBL {} Ljung, G.M.%
\end{APACrefauthors}%
\unskip\
\newblock
\APACrefYear{2016}.
\newblock
\APACrefbtitle {{Time series analysis: forecasting and control}} {{Time series analysis: forecasting and control}}\ (\PrintOrdinal{5}\ \BEd).
\newblock
\APACaddressPublisher{}{John Wiley {\&} Sons, Inc.}
\PrintBackRefs{\CurrentBib}

\bibitem [\protect \citeauthoryear {%
Breunig%
, Kriegel%
, Ng%
\BCBL {}\ \BBA {} Sander%
}{%
Breunig%
\ \protect \BOthers {.}}{%
{\protect \APACyear {2000}}%
}]{%
Breunig2000LOF:Outliers}
\APACinsertmetastar {%
Breunig2000LOF:Outliers}%
\begin{APACrefauthors}%
Breunig, M.M.%
, Kriegel, H\BHBI P.%
, Ng, R.T.%
\BCBL {} Sander, J.%
\end{APACrefauthors}%
\unskip\
\newblock
\APACrefYearMonthDay{2000}{5}{}.
\newblock
{\BBOQ}\APACrefatitle {{LOF: Identifying Density-Based Local Outliers}} {{LOF: Identifying Density-Based Local Outliers}}.{\BBCQ}
\newblock
 \APACrefbtitle {Proceedings of the 2000 ACM SIGMOD international conference on Management of data} {Proceedings of the 2000 acm sigmod international conference on management of data}\ (\BPGS\ 93--104).
\newblock
\APACaddressPublisher{New York, NY, USA}{ACM}.
\PrintBackRefs{\CurrentBib}

\bibitem [\protect \citeauthoryear {%
Campello%
, Moulavi%
, Zimek%
\BCBL {}\ \BBA {} Sander%
}{%
Campello%
\ \protect \BOthers {.}}{%
{\protect \APACyear {2015}}%
}]{%
Campello2015HierarchicalDetection}
\APACinsertmetastar {%
Campello2015HierarchicalDetection}%
\begin{APACrefauthors}%
Campello, R.J.G.B.%
, Moulavi, D.%
, Zimek, A.%
\BCBL {} Sander, J.%
\end{APACrefauthors}%
\unskip\
\newblock
\APACrefYearMonthDay{2015}{7}{}.
\newblock
{\BBOQ}\APACrefatitle {{Hierarchical Density Estimates for Data Clustering, Visualization, and Outlier Detection}} {{Hierarchical Density Estimates for Data Clustering, Visualization, and Outlier Detection}}.{\BBCQ}
\newblock
\APACjournalVolNumPages{ACM Transactions on Knowledge Discovery from Data}{10}{1}{1--51,}
\newblock
\begin{APACrefDOI} \doi{10.1145/2733381} \end{APACrefDOI}
\newblock

\newblock

\PrintBackRefs{\CurrentBib}

\bibitem [\protect \citeauthoryear {%
Chang%
, Terzis%
\BCBL {}\ \BBA {} Bonnet%
}{%
Chang%
\ \protect \BOthers {.}}{%
{\protect \APACyear {2009}}%
}]{%
Chang2009Mote-BasedNetworks}
\APACinsertmetastar {%
Chang2009Mote-BasedNetworks}%
\begin{APACrefauthors}%
Chang, M.%
, Terzis, A.%
\BCBL {} Bonnet, P.%
\end{APACrefauthors}%
\unskip\
\newblock
\APACrefYearMonthDay{2009}{}{}.
\newblock
{\BBOQ}\APACrefatitle {{Mote-Based Online Anomaly Detection Using Echo State Networks}} {{Mote-Based Online Anomaly Detection Using Echo State Networks}}.{\BBCQ}
\newblock
 \APACrefbtitle {Distributed Computing in Sensor Systems} {Distributed computing in sensor systems}\ (\BPGS\ 72--86).
\PrintBackRefs{\CurrentBib}

\bibitem [\protect \citeauthoryear {%
Chen%
\ \BBA {} Wu%
}{%
Chen%
\ \BBA {} Wu%
}{%
{\protect \APACyear {2025}}%
}]{%
Chen2025BayesianSeries}
\APACinsertmetastar {%
Chen2025BayesianSeries}%
\begin{APACrefauthors}%
Chen, X.%
\BCBT {}\ \BBA {} Wu, W.%
\end{APACrefauthors}%
\unskip\
\newblock
\APACrefYearMonthDay{2025}{8}{}.
\newblock
{\BBOQ}\APACrefatitle {{Bayesian online collective anomaly and change point detection in fine-grained time series}} {{Bayesian online collective anomaly and change point detection in fine-grained time series}}.{\BBCQ}
\newblock
\APACjournalVolNumPages{arXiv}{}{}{,}
\newblock

\newblock

\PrintBackRefs{\CurrentBib}

\bibitem [\protect \citeauthoryear {%
Cho%
\ \BBA {} Fryzlewicz%
}{%
Cho%
\ \BBA {} Fryzlewicz%
}{%
{\protect \APACyear {2015}}%
}]{%
Cho2015Multiple-Change-PointSegmentation}
\APACinsertmetastar {%
Cho2015Multiple-Change-PointSegmentation}%
\begin{APACrefauthors}%
Cho, H.%
\BCBT {}\ \BBA {} Fryzlewicz, P.%
\end{APACrefauthors}%
\unskip\
\newblock
\APACrefYearMonthDay{2015}{3}{}.
\newblock
{\BBOQ}\APACrefatitle {{Multiple-Change-Point Detection for High Dimensional Time Series via Sparsified Binary Segmentation}} {{Multiple-Change-Point Detection for High Dimensional Time Series via Sparsified Binary Segmentation}}.{\BBCQ}
\newblock
\APACjournalVolNumPages{Journal of the Royal Statistical Society Series B: Statistical Methodology}{77}{2}{475--507,}
\newblock
\begin{APACrefDOI} \doi{10.1111/rssb.12079} \end{APACrefDOI}
\newblock

\newblock

\PrintBackRefs{\CurrentBib}

\bibitem [\protect \citeauthoryear {%
Chu%
, Hornik%
\BCBL {}\ \BBA {} Kaun%
}{%
Chu%
\ \protect \BOthers {.}}{%
{\protect \APACyear {1995}}%
}]{%
Chu1995MOSUMConstancy}
\APACinsertmetastar {%
Chu1995MOSUMConstancy}%
\begin{APACrefauthors}%
Chu, C\BHBI S.J.%
, Hornik, K.%
\BCBL {} Kaun, C\BHBI M.%
\end{APACrefauthors}%
\unskip\
\newblock
\APACrefYearMonthDay{1995}{}{}.
\newblock
{\BBOQ}\APACrefatitle {{MOSUM tests for parameter constancy}} {{MOSUM tests for parameter constancy}}.{\BBCQ}
\newblock
\APACjournalVolNumPages{Biometrika}{82}{3}{603--617,}
\newblock
\begin{APACrefDOI} \doi{10.1093/biomet/82.3.603} \end{APACrefDOI}
\newblock

\newblock

\PrintBackRefs{\CurrentBib}

\bibitem [\protect \citeauthoryear {%
De~Graaf%
}{%
De~Graaf%
}{%
{\protect \APACyear {2019}}%
}]{%
DeGraaf2019InTechniques}
\APACinsertmetastar {%
DeGraaf2019InTechniques}%
\begin{APACrefauthors}%
De~Graaf, R.A.%
\end{APACrefauthors}%
\unskip\
\newblock
\APACrefYear{2019}.
\newblock
\APACrefbtitle {{In vivo NMR spectroscopy: principles and techniques}} {{In vivo NMR spectroscopy: principles and techniques}}\ (\PrintOrdinal{3rd ed}\ \BEd).
\newblock
\APACaddressPublisher{Hoboken, NJ}{John Wiley {\&} Sons, Inc}.
\PrintBackRefs{\CurrentBib}

\bibitem [\protect \citeauthoryear {%
Didi%
, Gafni%
\BCBL {}\ \BBA {} Cohen%
}{%
Didi%
\ \protect \BOthers {.}}{%
{\protect \APACyear {2024}}%
}]{%
Didi2024AsymptoticallyModel}
\APACinsertmetastar {%
Didi2024AsymptoticallyModel}%
\begin{APACrefauthors}%
Didi, L.L.%
, Gafni, T.%
\BCBL {} Cohen, K.%
\end{APACrefauthors}%
\unskip\
\newblock
\APACrefYearMonthDay{2024}{12}{}.
\newblock
{\BBOQ}\APACrefatitle {{Asymptotically Optimal Search for a Change Point Anomaly under a Composite Hypothesis Model}} {{Asymptotically Optimal Search for a Change Point Anomaly under a Composite Hypothesis Model}}.{\BBCQ}
\newblock
\APACjournalVolNumPages{arXiv}{}{}{,}
\newblock

\newblock

\PrintBackRefs{\CurrentBib}

\bibitem [\protect \citeauthoryear {%
Duong%
, Le%
\BCBL {}\ \BBA {} Hoang%
}{%
Duong%
\ \protect \BOthers {.}}{%
{\protect \APACyear {2023}}%
}]{%
Duong2023DeepSurvey}
\APACinsertmetastar {%
Duong2023DeepSurvey}%
\begin{APACrefauthors}%
Duong, H\BHBI T.%
, Le, V\BHBI T.%
\BCBL {} Hoang, V.T.%
\end{APACrefauthors}%
\unskip\
\newblock
\APACrefYearMonthDay{2023}{5}{}.
\newblock
{\BBOQ}\APACrefatitle {{Deep Learning-Based Anomaly Detection in Video Surveillance: A Survey}} {{Deep Learning-Based Anomaly Detection in Video Surveillance: A Survey}}.{\BBCQ}
\newblock
\APACjournalVolNumPages{Sensors}{23}{11}{5024,}
\newblock
\begin{APACrefDOI} \doi{10.3390/s23115024} \end{APACrefDOI}
\newblock

\newblock

\PrintBackRefs{\CurrentBib}

\bibitem [\protect \citeauthoryear {%
Elnaggar%
, Chakrabarty%
\BCBL {}\ \BBA {} Tahoori%
}{%
Elnaggar%
\ \protect \BOthers {.}}{%
{\protect \APACyear {2019}}%
}]{%
Elnaggar2019HardwareTechniques}
\APACinsertmetastar {%
Elnaggar2019HardwareTechniques}%
\begin{APACrefauthors}%
Elnaggar, R.%
, Chakrabarty, K.%
\BCBL {} Tahoori, M.B.%
\end{APACrefauthors}%
\unskip\
\newblock
\APACrefYearMonthDay{2019}{12}{}.
\newblock
{\BBOQ}\APACrefatitle {{Hardware Trojan Detection Using Changepoint-Based Anomaly Detection Techniques}} {{Hardware Trojan Detection Using Changepoint-Based Anomaly Detection Techniques}}.{\BBCQ}
\newblock
\APACjournalVolNumPages{IEEE Transactions on Very Large Scale Integration (VLSI) Systems}{27}{12}{2706--2719,}
\newblock
\begin{APACrefDOI} \doi{10.1109/TVLSI.2019.2925807} \end{APACrefDOI}
\newblock

\newblock

\PrintBackRefs{\CurrentBib}

\bibitem [\protect \citeauthoryear {%
Fahim%
\ \BBA {} Sillitti%
}{%
Fahim%
\ \BBA {} Sillitti%
}{%
{\protect \APACyear {2019}}%
}]{%
Fahim2019AnomalyReview}
\APACinsertmetastar {%
Fahim2019AnomalyReview}%
\begin{APACrefauthors}%
Fahim, M.%
\BCBT {}\ \BBA {} Sillitti, A.%
\end{APACrefauthors}%
\unskip\
\newblock
\APACrefYearMonthDay{2019}{}{}.
\newblock
{\BBOQ}\APACrefatitle {{Anomaly Detection, Analysis and Prediction Techniques in IoT Environment: A Systematic Literature Review}} {{Anomaly Detection, Analysis and Prediction Techniques in IoT Environment: A Systematic Literature Review}}.{\BBCQ}
\newblock
\APACjournalVolNumPages{IEEE Access}{7}{}{81664--81681,}
\newblock
\begin{APACrefDOI} \doi{10.1109/ACCESS.2019.2921912} \end{APACrefDOI}
\newblock

\newblock

\PrintBackRefs{\CurrentBib}

\bibitem [\protect \citeauthoryear {%
Fisch%
, Eckley%
\BCBL {}\ \BBA {} Fearnhead%
}{%
Fisch%
\ \protect \BOthers {.}}{%
{\protect \APACyear {2022}}%
}]{%
Fisch2022AAnomalies}
\APACinsertmetastar {%
Fisch2022AAnomalies}%
\begin{APACrefauthors}%
Fisch, A.T.%
, Eckley, I.A.%
\BCBL {} Fearnhead, P.%
\end{APACrefauthors}%
\unskip\
\newblock
\APACrefYearMonthDay{2022}{8}{}.
\newblock
{\BBOQ}\APACrefatitle {{A linear time method for the detection of collective and point anomalies}} {{A linear time method for the detection of collective and point anomalies}}.{\BBCQ}
\newblock
\APACjournalVolNumPages{Statistical Analysis and Data Mining}{15}{4}{494--508,}
\newblock
\begin{APACrefDOI} \doi{10.1002/sam.11586} \end{APACrefDOI}
\newblock

\newblock

\PrintBackRefs{\CurrentBib}

\bibitem [\protect \citeauthoryear {%
Fryzlewicz%
}{%
Fryzlewicz%
}{%
{\protect \APACyear {2014}}%
}]{%
Fryzlewicz2014WildDetection}
\APACinsertmetastar {%
Fryzlewicz2014WildDetection}%
\begin{APACrefauthors}%
Fryzlewicz, P.%
\end{APACrefauthors}%
\unskip\
\newblock
\APACrefYearMonthDay{2014}{12}{}.
\newblock
{\BBOQ}\APACrefatitle {{Wild binary segmentation for multiple change-point detection}} {{Wild binary segmentation for multiple change-point detection}}.{\BBCQ}
\newblock
\APACjournalVolNumPages{Annals of Statistics}{42}{6}{2243--2281,}
\newblock
\begin{APACrefDOI} \doi{10.1214/14-AOS1245} \end{APACrefDOI}
\newblock

\newblock

\PrintBackRefs{\CurrentBib}

\bibitem [\protect \citeauthoryear {%
Gebski%
\ \BBA {} Wong%
}{%
Gebski%
\ \BBA {} Wong%
}{%
{\protect \APACyear {2007}}%
}]{%
Gebski2007AnDetection}
\APACinsertmetastar {%
Gebski2007AnDetection}%
\begin{APACrefauthors}%
Gebski, M.%
\BCBT {}\ \BBA {} Wong, R.K.%
\end{APACrefauthors}%
\unskip\
\newblock
\APACrefYearMonthDay{2007}{}{}.
\newblock
{\BBOQ}\APACrefatitle {{An Efficient Histogram Method for Outlier Detection}} {{An Efficient Histogram Method for Outlier Detection}}.{\BBCQ}
\newblock
 \APACrefbtitle {Advances in Databases: Concepts, Systems and Applications} {Advances in databases: Concepts, systems and applications}\ (\BPGS\ 176--187).
\newblock
\APACaddressPublisher{Berlin, Heidelberg}{Springer Berlin Heidelberg}.
\PrintBackRefs{\CurrentBib}

\bibitem [\protect \citeauthoryear {%
Geiger%
, Liu%
, Alnegheimish%
, Cuesta-Infante%
\BCBL {}\ \BBA {} Veeramachaneni%
}{%
Geiger%
\ \protect \BOthers {.}}{%
{\protect \APACyear {2020}}%
}]{%
Geiger2020TadGAN:Networks}
\APACinsertmetastar {%
Geiger2020TadGAN:Networks}%
\begin{APACrefauthors}%
Geiger, A.%
, Liu, D.%
, Alnegheimish, S.%
, Cuesta-Infante, A.%
\BCBL {} Veeramachaneni, K.%
\end{APACrefauthors}%
\unskip\
\newblock
\APACrefYearMonthDay{2020}{12}{}.
\newblock
{\BBOQ}\APACrefatitle {{TadGAN: Time Series Anomaly Detection Using Generative Adversarial Networks}} {{TadGAN: Time Series Anomaly Detection Using Generative Adversarial Networks}}.{\BBCQ}
\newblock
 \APACrefbtitle {Proceedings - 2020 IEEE International Conference on Big Data, Big Data 2020} {Proceedings - 2020 ieee international conference on big data, big data 2020}\ (\BPGS\ 33--43).
\newblock
\APACaddressPublisher{}{Institute of Electrical and Electronics Engineers Inc.}
\PrintBackRefs{\CurrentBib}

\bibitem [\protect \citeauthoryear {%
Govindaraju%
\ \protect \BOthers {.}}{%
Govindaraju%
\ \protect \BOthers {.}}{%
{\protect \APACyear {2004}}%
}]{%
Govindaraju2004VolumetricInjury.}
\APACinsertmetastar {%
Govindaraju2004VolumetricInjury.}%
\begin{APACrefauthors}%
Govindaraju, V.%
, Gauger, G.E.%
, Manley, G.T.%
, Ebel, A.%
, Meeker, M.%
\BCBL {} Maudsley, A.A.%
\end{APACrefauthors}%
\unskip\
\newblock
\APACrefYearMonthDay{2004}{5}{}.
\newblock
{\BBOQ}\APACrefatitle {{Volumetric proton spectroscopic imaging of mild traumatic brain injury.}} {{Volumetric proton spectroscopic imaging of mild traumatic brain injury.}}{\BBCQ}
\newblock
\APACjournalVolNumPages{AJNR. American journal of neuroradiology}{25}{5}{730--7,}
\newblock

\newblock

\PrintBackRefs{\CurrentBib}

\bibitem [\protect \citeauthoryear {%
Grubbs%
}{%
Grubbs%
}{%
{\protect \APACyear {1969}}%
}]{%
Grubbs1969ProceduresSamples}
\APACinsertmetastar {%
Grubbs1969ProceduresSamples}%
\begin{APACrefauthors}%
Grubbs, F.E.%
\end{APACrefauthors}%
\unskip\
\newblock
\APACrefYearMonthDay{1969}{2}{}.
\newblock
{\BBOQ}\APACrefatitle {{Procedures for Detecting Outlying Observations in Samples}} {{Procedures for Detecting Outlying Observations in Samples}}.{\BBCQ}
\newblock
\APACjournalVolNumPages{Technometrics}{11}{1}{1--21,}
\newblock
\begin{APACrefDOI} \doi{10.1080/00401706.1969.10490657} \end{APACrefDOI}
\newblock

\newblock

\PrintBackRefs{\CurrentBib}

\bibitem [\protect \citeauthoryear {%
Gurbani%
\ \protect \BOthers {.}}{%
Gurbani%
\ \protect \BOthers {.}}{%
{\protect \APACyear {2018}}%
}]{%
Gurbani2018AMRI}
\APACinsertmetastar {%
Gurbani2018AMRI}%
\begin{APACrefauthors}%
Gurbani, S.S.%
, Schreibmann, E.%
, Maudsley, A.A.%
, Cordova, J.S.%
, Soher, B.J.%
, Poptani, H.%
\BDBL {}Cooper, L.A.D.%
\end{APACrefauthors}%
\unskip\
\newblock
\APACrefYearMonthDay{2018}{11}{}.
\newblock
{\BBOQ}\APACrefatitle {{A convolutional neural network to filter artifacts in spectroscopic MRI}} {{A convolutional neural network to filter artifacts in spectroscopic MRI}}.{\BBCQ}
\newblock
\APACjournalVolNumPages{Magnetic Resonance in Medicine}{80}{5}{1765--1775,}
\newblock
\begin{APACrefDOI} \doi{10.1002/mrm.27166} \end{APACrefDOI}
\newblock

\newblock

\PrintBackRefs{\CurrentBib}

\bibitem [\protect \citeauthoryear {%
Harchaoui%
, Bach%
\BCBL {}\ \BBA {} Moulines%
}{%
Harchaoui%
\ \protect \BOthers {.}}{%
{\protect \APACyear {2008}}%
}]{%
Harchaoui2008KernelAnalysis}
\APACinsertmetastar {%
Harchaoui2008KernelAnalysis}%
\begin{APACrefauthors}%
Harchaoui, Z.%
, Bach, F.%
\BCBL {} Moulines, E.%
\end{APACrefauthors}%
\unskip\
\newblock
\APACrefYearMonthDay{2008}{}{}.
\newblock
{\BBOQ}\APACrefatitle {{Kernel Change-point Analysis}} {{Kernel Change-point Analysis}}.{\BBCQ}
\newblock
 \APACrefbtitle {Advances in Neural Information Processing Systems 21 (NIPS 2008).} {Advances in neural information processing systems 21 (nips 2008).}
\PrintBackRefs{\CurrentBib}

\bibitem [\protect \citeauthoryear {%
He%
, Zhang%
, Ren%
\BCBL {}\ \BBA {} Sun%
}{%
He%
\ \protect \BOthers {.}}{%
{\protect \APACyear {2016}}%
}]{%
He2016DeepRecognition}
\APACinsertmetastar {%
He2016DeepRecognition}%
\begin{APACrefauthors}%
He, K.%
, Zhang, X.%
, Ren, S.%
\BCBL {} Sun, J.%
\end{APACrefauthors}%
\unskip\
\newblock
\APACrefYearMonthDay{2016}{6}{}.
\newblock
{\BBOQ}\APACrefatitle {{Deep Residual Learning for Image Recognition}} {{Deep Residual Learning for Image Recognition}}.{\BBCQ}
\newblock
 \APACrefbtitle {2016 IEEE Conference on Computer Vision and Pattern Recognition (CVPR)} {2016 ieee conference on computer vision and pattern recognition (cvpr)}\ (\BPGS\ 770--778).
\newblock
\APACaddressPublisher{}{IEEE}.
\PrintBackRefs{\CurrentBib}

\bibitem [\protect \citeauthoryear {%
Hilal%
, Gadsden%
\BCBL {}\ \BBA {} Yawney%
}{%
Hilal%
\ \protect \BOthers {.}}{%
{\protect \APACyear {2022}}%
}]{%
Hilal2022FinancialAdvances}
\APACinsertmetastar {%
Hilal2022FinancialAdvances}%
\begin{APACrefauthors}%
Hilal, W.%
, Gadsden, S.A.%
\BCBL {} Yawney, J.%
\end{APACrefauthors}%
\unskip\
\newblock
\APACrefYearMonthDay{2022}{5}{}.
\newblock
{\BBOQ}\APACrefatitle {{Financial Fraud: A Review of Anomaly Detection Techniques and Recent Advances}} {{Financial Fraud: A Review of Anomaly Detection Techniques and Recent Advances}}.{\BBCQ}
\newblock
\APACjournalVolNumPages{Expert Systems with Applications}{193}{}{116429,}
\newblock
\begin{APACrefDOI} \doi{10.1016/j.eswa.2021.116429} \end{APACrefDOI}
\newblock

\newblock

\PrintBackRefs{\CurrentBib}

\bibitem [\protect \citeauthoryear {%
Huijben%
, Kool%
, Paulus%
\BCBL {}\ \BBA {} van Sloun%
}{%
Huijben%
\ \protect \BOthers {.}}{%
{\protect \APACyear {2023}}%
}]{%
Huijben2023ALearning}
\APACinsertmetastar {%
Huijben2023ALearning}%
\begin{APACrefauthors}%
Huijben, I.A.M.%
, Kool, W.%
, Paulus, M.B.%
\BCBL {} van Sloun, R.J.G.%
\end{APACrefauthors}%
\unskip\
\newblock
\APACrefYearMonthDay{2023}{2}{}.
\newblock
{\BBOQ}\APACrefatitle {{A Review of the Gumbel-max Trick and its Extensions for Discrete Stochasticity in Machine Learning}} {{A Review of the Gumbel-max Trick and its Extensions for Discrete Stochasticity in Machine Learning}}.{\BBCQ}
\newblock
\APACjournalVolNumPages{IEEE Transactions on Pattern Analysis and Machine Intelligence}{45}{2}{1353--1371,}
\newblock
\begin{APACrefDOI} \doi{10.1109/TPAMI.2022.3157042} \end{APACrefDOI}
\newblock
\begin{APACrefURL} {https://ieeexplore.ieee.org/document/9729603/} \end{APACrefURL}
\newblock

\newblock

\PrintBackRefs{\CurrentBib}

\bibitem [\protect \citeauthoryear {%
E.~Jang%
, Gu%
\BCBL {}\ \BBA {} Poole%
}{%
E.~Jang%
\ \protect \BOthers {.}}{%
{\protect \APACyear {2017}}%
}]{%
Jang2017CategoricalGumbel-Softmax}
\APACinsertmetastar {%
Jang2017CategoricalGumbel-Softmax}%
\begin{APACrefauthors}%
Jang, E.%
, Gu, S.%
\BCBL {} Poole, B.%
\end{APACrefauthors}%
\unskip\
\newblock
\APACrefYearMonthDay{2017}{11}{}.
\newblock
{\BBOQ}\APACrefatitle {{Categorical Reparameterization with Gumbel-Softmax}} {{Categorical Reparameterization with Gumbel-Softmax}}.{\BBCQ}
\newblock
 \APACrefbtitle {International Conference on Learning Representations.} {International conference on learning representations.}
\PrintBackRefs{\CurrentBib}

\bibitem [\protect \citeauthoryear {%
J.~Jang%
, Lee%
, Park%
\BCBL {}\ \BBA {} Kim%
}{%
J.~Jang%
\ \protect \BOthers {.}}{%
{\protect \APACyear {2021}}%
}]{%
Jang2021UnsupervisedBrain}
\APACinsertmetastar {%
Jang2021UnsupervisedBrain}%
\begin{APACrefauthors}%
Jang, J.%
, Lee, H.H.%
, Park, J\BHBI A.%
\BCBL {} Kim, H.%
\end{APACrefauthors}%
\unskip\
\newblock
\APACrefYearMonthDay{2021}{4}{}.
\newblock
{\BBOQ}\APACrefatitle {{Unsupervised anomaly detection using generative adversarial networks in 1H-MRS of the brain}} {{Unsupervised anomaly detection using generative adversarial networks in 1H-MRS of the brain}}.{\BBCQ}
\newblock
\APACjournalVolNumPages{Journal of Magnetic Resonance}{325}{}{106936,}
\newblock
\begin{APACrefDOI} \doi{10.1016/j.jmr.2021.106936} \end{APACrefDOI}
\newblock

\newblock

\PrintBackRefs{\CurrentBib}

\bibitem [\protect \citeauthoryear {%
M.~Jiang%
, Tseng%
\BCBL {}\ \BBA {} Su%
}{%
M.~Jiang%
\ \protect \BOthers {.}}{%
{\protect \APACyear {2001}}%
}]{%
Jiang2001Two-phaseDetection}
\APACinsertmetastar {%
Jiang2001Two-phaseDetection}%
\begin{APACrefauthors}%
Jiang, M.%
, Tseng, S.%
\BCBL {} Su, C.%
\end{APACrefauthors}%
\unskip\
\newblock
\APACrefYearMonthDay{2001}{5}{}.
\newblock
{\BBOQ}\APACrefatitle {{Two-phase clustering process for outliers detection}} {{Two-phase clustering process for outliers detection}}.{\BBCQ}
\newblock
\APACjournalVolNumPages{Pattern Recognition Letters}{22}{6-7}{691--700,}
\newblock
\begin{APACrefDOI} \doi{10.1016/S0167-8655(00)00131-8} \end{APACrefDOI}
\newblock

\newblock

\PrintBackRefs{\CurrentBib}

\bibitem [\protect \citeauthoryear {%
S\BHBI y.~Jiang%
\ \BBA {} An%
}{%
S\BHBI y.~Jiang%
\ \BBA {} An%
}{%
{\protect \APACyear {2008}}%
}]{%
Jiang2008Clustering-BasedMethod}
\APACinsertmetastar {%
Jiang2008Clustering-BasedMethod}%
\begin{APACrefauthors}%
Jiang, S\BHBI y.%
\BCBT {}\ \BBA {} An, Q\BHBI b.%
\end{APACrefauthors}%
\unskip\
\newblock
\APACrefYearMonthDay{2008}{10}{}.
\newblock
{\BBOQ}\APACrefatitle {{Clustering-Based Outlier Detection Method}} {{Clustering-Based Outlier Detection Method}}.{\BBCQ}
\newblock
 \APACrefbtitle {2008 Fifth International Conference on Fuzzy Systems and Knowledge Discovery} {2008 fifth international conference on fuzzy systems and knowledge discovery}\ (\BVOL~2, \BPGS\ 429--433).
\newblock
\APACaddressPublisher{}{IEEE}.
\PrintBackRefs{\CurrentBib}

\bibitem [\protect \citeauthoryear {%
Killick%
, Fearnhead%
\BCBL {}\ \BBA {} Eckley%
}{%
Killick%
\ \protect \BOthers {.}}{%
{\protect \APACyear {2012}}%
}]{%
Killick2012OptimalCost}
\APACinsertmetastar {%
Killick2012OptimalCost}%
\begin{APACrefauthors}%
Killick, R.%
, Fearnhead, P.%
\BCBL {} Eckley, I.A.%
\end{APACrefauthors}%
\unskip\
\newblock
\APACrefYearMonthDay{2012}{}{}.
\newblock
{\BBOQ}\APACrefatitle {{Optimal detection of changepoints with a linear computational cost}} {{Optimal detection of changepoints with a linear computational cost}}.{\BBCQ}
\newblock
\APACjournalVolNumPages{Journal of the American Statistical Association}{107}{500}{1590--1598,}
\newblock
\begin{APACrefDOI} \doi{10.1080/01621459.2012.737745} \end{APACrefDOI}
\newblock

\newblock

\PrintBackRefs{\CurrentBib}

\bibitem [\protect \citeauthoryear {%
Knorr%
, Ng%
\BCBL {}\ \BBA {} Tucakov%
}{%
Knorr%
\ \protect \BOthers {.}}{%
{\protect \APACyear {2000}}%
}]{%
Knorr2000Distance-basedApplications}
\APACinsertmetastar {%
Knorr2000Distance-basedApplications}%
\begin{APACrefauthors}%
Knorr, E.M.%
, Ng, R.T.%
\BCBL {} Tucakov, V.%
\end{APACrefauthors}%
\unskip\
\newblock
\APACrefYearMonthDay{2000}{2}{}.
\newblock
{\BBOQ}\APACrefatitle {{Distance-based outliers: algorithms and applications}} {{Distance-based outliers: algorithms and applications}}.{\BBCQ}
\newblock
\APACjournalVolNumPages{The VLDB Journal The International Journal on Very Large Data Bases}{8}{3-4}{237--253,}
\newblock
\begin{APACrefDOI} \doi{10.1007/s007780050006} \end{APACrefDOI}
\newblock

\newblock

\PrintBackRefs{\CurrentBib}

\bibitem [\protect \citeauthoryear {%
Kreis%
}{%
Kreis%
}{%
{\protect \APACyear {2004}}%
}]{%
Kreis2004IssuesArtifacts}
\APACinsertmetastar {%
Kreis2004IssuesArtifacts}%
\begin{APACrefauthors}%
Kreis, R.%
\end{APACrefauthors}%
\unskip\
\newblock
\APACrefYearMonthDay{2004}{10}{}.
\newblock
{\BBOQ}\APACrefatitle {{Issues of spectral quality in clinical 1H‐magnetic resonance spectroscopy and a gallery of artifacts}} {{Issues of spectral quality in clinical 1H‐magnetic resonance spectroscopy and a gallery of artifacts}}.{\BBCQ}
\newblock
\APACjournalVolNumPages{NMR in Biomedicine}{17}{6}{361--381,}
\newblock
\begin{APACrefDOI} \doi{10.1002/nbm.891} \end{APACrefDOI}
\newblock

\newblock

\PrintBackRefs{\CurrentBib}

\bibitem [\protect \citeauthoryear {%
Kreis%
\ \protect \BOthers {.}}{%
Kreis%
\ \protect \BOthers {.}}{%
{\protect \APACyear {2021}}%
}]{%
Kreis2021TerminologyRecommendations}
\APACinsertmetastar {%
Kreis2021TerminologyRecommendations}%
\begin{APACrefauthors}%
Kreis, R.%
, Boer, V.%
, Choi, I.%
, Cudalbu, C.%
, de Graaf, R.A.%
, Gasparovic, C.%
\BDBL {}Bogner, W.%
\end{APACrefauthors}%
\unskip\
\newblock
\APACrefYearMonthDay{2021}{5}{}.
\newblock
{\BBOQ}\APACrefatitle {{Terminology and concepts for the characterization of in vivo MR spectroscopy methods and MR spectra: Background and experts' consensus recommendations}} {{Terminology and concepts for the characterization of in vivo MR spectroscopy methods and MR spectra: Background and experts' consensus recommendations}}.{\BBCQ}
\newblock
\APACjournalVolNumPages{NMR in Biomedicine}{34}{5}{,}
\newblock
\begin{APACrefDOI} \doi{10.1002/nbm.4347} \end{APACrefDOI}
\newblock

\newblock

\PrintBackRefs{\CurrentBib}

\bibitem [\protect \citeauthoryear {%
Kriegel%
, Kr{\"{o}}ger%
, Schubert%
\BCBL {}\ \BBA {} Zimek%
}{%
Kriegel%
\ \protect \BOthers {.}}{%
{\protect \APACyear {2009}}%
}]{%
Kriegel2009OutlierData}
\APACinsertmetastar {%
Kriegel2009OutlierData}%
\begin{APACrefauthors}%
Kriegel, H\BHBI P.%
, Kr{\"{o}}ger, P.%
, Schubert, E.%
\BCBL {} Zimek, A.%
\end{APACrefauthors}%
\unskip\
\newblock
\APACrefYearMonthDay{2009}{}{}.
\newblock
{\BBOQ}\APACrefatitle {{Outlier Detection in Axis-Parallel Subspaces of High Dimensional Data}} {{Outlier Detection in Axis-Parallel Subspaces of High Dimensional Data}}.{\BBCQ}
\newblock
 (\BPGS\ 831--838).
\PrintBackRefs{\CurrentBib}

\bibitem [\protect \citeauthoryear {%
Kyathanahally%
\ \protect \BOthers {.}}{%
Kyathanahally%
\ \protect \BOthers {.}}{%
{\protect \APACyear {2018}}%
}]{%
Kyathanahally2018QualityTools}
\APACinsertmetastar {%
Kyathanahally2018QualityTools}%
\begin{APACrefauthors}%
Kyathanahally, S.P.%
, Mocioiu, V.%
, Pedrosa~de Barros, N.%
, Slotboom, J.%
, Wright, A.J.%
, Juli{\`{a}}‐Sap{\'{e}}, M.%
\BDBL {}Kreis, R.%
\end{APACrefauthors}%
\unskip\
\newblock
\APACrefYearMonthDay{2018}{5}{}.
\newblock
{\BBOQ}\APACrefatitle {{Quality of clinical brain tumor MR spectra judged by humans and machine learning tools}} {{Quality of clinical brain tumor MR spectra judged by humans and machine learning tools}}.{\BBCQ}
\newblock
\APACjournalVolNumPages{Magnetic Resonance in Medicine}{79}{5}{2500--2510,}
\newblock
\begin{APACrefDOI} \doi{10.1002/mrm.26948} \end{APACrefDOI}
\newblock

\newblock

\PrintBackRefs{\CurrentBib}

\bibitem [\protect \citeauthoryear {%
Landheer%
\ \BBA {} Juchem%
}{%
Landheer%
\ \BBA {} Juchem%
}{%
{\protect \APACyear {2021}}%
}]{%
Landheer2021AreSpectroscopy}
\APACinsertmetastar {%
Landheer2021AreSpectroscopy}%
\begin{APACrefauthors}%
Landheer, K.%
\BCBT {}\ \BBA {} Juchem, C.%
\end{APACrefauthors}%
\unskip\
\newblock
\APACrefYearMonthDay{2021}{7}{}.
\newblock
{\BBOQ}\APACrefatitle {{Are Cram{\'{e}}r‐Rao lower bounds an accurate estimate for standard deviations in in vivo magnetic resonance spectroscopy?}} {{Are Cram{\'{e}}r‐Rao lower bounds an accurate estimate for standard deviations in in vivo magnetic resonance spectroscopy?}}{\BBCQ}
\newblock
\APACjournalVolNumPages{NMR in Biomedicine}{34}{7}{,}
\newblock
\begin{APACrefDOI} \doi{10.1002/nbm.4521} \end{APACrefDOI}
\newblock

\newblock

\PrintBackRefs{\CurrentBib}

\bibitem [\protect \citeauthoryear {%
Lecun%
, Bottou%
, Bengio%
\BCBL {}\ \BBA {} Haffner%
}{%
Lecun%
\ \protect \BOthers {.}}{%
{\protect \APACyear {1998}}%
}]{%
Lecun1998Gradient-BasedRecognition}
\APACinsertmetastar {%
Lecun1998Gradient-BasedRecognition}%
\begin{APACrefauthors}%
Lecun, Y.%
, Bottou, L.%
, Bengio, Y.%
\BCBL {} Haffner, P.%
\end{APACrefauthors}%
\unskip\
\newblock
\APACrefYearMonthDay{1998}{}{}.
\newblock
{\BBOQ}\APACrefatitle {{Gradient-based learning applied to document recognition}} {{Gradient-based learning applied to document recognition}}.{\BBCQ}
\newblock
\APACjournalVolNumPages{Proceedings of the IEEE}{86}{11}{2278--2324,}
\newblock
\begin{APACrefDOI} \doi{10.1109/5.726791} \end{APACrefDOI}
\newblock

\newblock

\PrintBackRefs{\CurrentBib}

\bibitem [\protect \citeauthoryear {%
Li%
, Fearnhead%
, Fryzlewicz%
\BCBL {}\ \BBA {} Wang%
}{%
Li%
\ \protect \BOthers {.}}{%
{\protect \APACyear {2024}}%
}]{%
Li2024AutomaticLearning}
\APACinsertmetastar {%
Li2024AutomaticLearning}%
\begin{APACrefauthors}%
Li, J.%
, Fearnhead, P.%
, Fryzlewicz, P.%
\BCBL {} Wang, T.%
\end{APACrefauthors}%
\unskip\
\newblock
\APACrefYearMonthDay{2024}{4}{}.
\newblock
{\BBOQ}\APACrefatitle {{Automatic change-point detection in time series via deep learning}} {{Automatic change-point detection in time series via deep learning}}.{\BBCQ}
\newblock
\APACjournalVolNumPages{Journal of the Royal Statistical Society Series B: Statistical Methodology}{86}{2}{273--285,}
\newblock
\begin{APACrefDOI} \doi{10.1093/jrsssb/qkae004} \end{APACrefDOI}
\newblock

\newblock

\PrintBackRefs{\CurrentBib}

\bibitem [\protect \citeauthoryear {%
Lin%
\ \protect \BOthers {.}}{%
Lin%
\ \protect \BOthers {.}}{%
{\protect \APACyear {2020}}%
}]{%
Lin2020AnomalyModel}
\APACinsertmetastar {%
Lin2020AnomalyModel}%
\begin{APACrefauthors}%
Lin, S.%
, Clark, R.%
, Birke, R.%
, Schonborn, S.%
, Trigoni, N.%
\BCBL {} Roberts, S.%
\end{APACrefauthors}%
\unskip\
\newblock
\APACrefYearMonthDay{2020}{5}{}.
\newblock
{\BBOQ}\APACrefatitle {{Anomaly Detection for Time Series Using VAE-LSTM Hybrid Model}} {{Anomaly Detection for Time Series Using VAE-LSTM Hybrid Model}}.{\BBCQ}
\newblock
 \APACrefbtitle {ICASSP 2020 - 2020 IEEE International Conference on Acoustics, Speech and Signal Processing (ICASSP)} {Icassp 2020 - 2020 ieee international conference on acoustics, speech and signal processing (icassp)}\ (\BPGS\ 4322--4326).
\newblock
\APACaddressPublisher{}{IEEE}.
\newblock
\begin{APACrefURL} {https://ieeexplore.ieee.org/document/9053558/} \end{APACrefURL}
\PrintBackRefs{\CurrentBib}

\bibitem [\protect \citeauthoryear {%
F.T.~Liu%
, Ting%
\BCBL {}\ \BBA {} Zhou%
}{%
F.T.~Liu%
\ \protect \BOthers {.}}{%
{\protect \APACyear {2012}}%
}]{%
Liu2012Isolation-BasedDetection}
\APACinsertmetastar {%
Liu2012Isolation-BasedDetection}%
\begin{APACrefauthors}%
Liu, F.T.%
, Ting, K.M.%
\BCBL {} Zhou, Z\BHBI H.%
\end{APACrefauthors}%
\unskip\
\newblock
\APACrefYearMonthDay{2012}{3}{}.
\newblock
{\BBOQ}\APACrefatitle {{Isolation-Based Anomaly Detection}} {{Isolation-Based Anomaly Detection}}.{\BBCQ}
\newblock
\APACjournalVolNumPages{ACM Transactions on Knowledge Discovery from Data}{6}{1}{1--39,}
\newblock
\begin{APACrefDOI} \doi{10.1145/2133360.2133363} \end{APACrefDOI}
\newblock

\newblock

\PrintBackRefs{\CurrentBib}

\bibitem [\protect \citeauthoryear {%
J.~Liu%
, Yang%
, Zhang%
, Gao%
\BCBL {}\ \BBA {} Li%
}{%
J.~Liu%
\ \protect \BOthers {.}}{%
{\protect \APACyear {2023}}%
}]{%
Liu2023AnomalyDrift}
\APACinsertmetastar {%
Liu2023AnomalyDrift}%
\begin{APACrefauthors}%
Liu, J.%
, Yang, D.%
, Zhang, K.%
, Gao, H.%
\BCBL {} Li, J.%
\end{APACrefauthors}%
\unskip\
\newblock
\APACrefYearMonthDay{2023}{9}{}.
\newblock
{\BBOQ}\APACrefatitle {{Anomaly and change point detection for time series with concept drift}} {{Anomaly and change point detection for time series with concept drift}}.{\BBCQ}
\newblock
\APACjournalVolNumPages{World Wide Web}{26}{5}{3229--3252,}
\newblock
\begin{APACrefDOI} \doi{10.1007/s11280-023-01181-z} \end{APACrefDOI}
\newblock

\newblock

\PrintBackRefs{\CurrentBib}

\bibitem [\protect \citeauthoryear {%
S.~Liu%
\ \protect \BOthers {.}}{%
S.~Liu%
\ \protect \BOthers {.}}{%
{\protect \APACyear {2024}}%
}]{%
Liu2024AnomalyLLM:Models}
\APACinsertmetastar {%
Liu2024AnomalyLLM:Models}%
\begin{APACrefauthors}%
Liu, S.%
, Yao, D.%
, Fang, L.%
, Li, Z.%
, Li, W.%
, Feng, K.%
\BDBL {}Bi, J.%
\end{APACrefauthors}%
\unskip\
\newblock
\APACrefYearMonthDay{2024}{}{}.
\newblock
{\BBOQ}\APACrefatitle {{AnomalyLLM: Few-Shot Anomaly Edge Detection for Dynamic Graphs Using Large Language Models}} {{AnomalyLLM: Few-Shot Anomaly Edge Detection for Dynamic Graphs Using Large Language Models}}.{\BBCQ}
\newblock
 \APACrefbtitle {Proceedings - IEEE International Conference on Data Mining, ICDM} {Proceedings - ieee international conference on data mining, icdm}\ (\BPGS\ 785--790).
\newblock
\APACaddressPublisher{}{Institute of Electrical and Electronics Engineers Inc.}
\PrintBackRefs{\CurrentBib}

\bibitem [\protect \citeauthoryear {%
Malhotra%
\ \protect \BOthers {.}}{%
Malhotra%
\ \protect \BOthers {.}}{%
{\protect \APACyear {2016}}%
}]{%
Malhotra2016}
\APACinsertmetastar {%
Malhotra2016}%
\begin{APACrefauthors}%
Malhotra, P.%
, Ramakrishnan, A.%
, Anand, G.%
, Vig, L.%
, Agarwal, P.%
\BCBL {} Shroff, G.%
\end{APACrefauthors}%
\unskip\
\newblock
\APACrefYearMonthDay{2016}{7}{}.
\newblock
{\BBOQ}\APACrefatitle {{LSTM-based Encoder-Decoder for Multi-sensor Anomaly Detection}} {{LSTM-based Encoder-Decoder for Multi-sensor Anomaly Detection}}.{\BBCQ}
\newblock
\APACjournalVolNumPages{arXiv}{}{}{,}
\newblock

\newblock

\PrintBackRefs{\CurrentBib}

\bibitem [\protect \citeauthoryear {%
Malhotra%
, Vig%
, Shroff%
\BCBL {}\ \BBA {} Agarwal%
}{%
Malhotra%
\ \protect \BOthers {.}}{%
{\protect \APACyear {2015}}%
}]{%
Malhotra2015LongSeries}
\APACinsertmetastar {%
Malhotra2015LongSeries}%
\begin{APACrefauthors}%
Malhotra, P.%
, Vig, L.%
, Shroff, G.%
\BCBL {} Agarwal, P.%
\end{APACrefauthors}%
\unskip\
\newblock
\APACrefYearMonthDay{2015}{}{}.
\newblock
{\BBOQ}\APACrefatitle {{Long short term memory networks for anomaly detection in time series}} {{Long short term memory networks for anomaly detection in time series}}.{\BBCQ}
\newblock
 \APACrefbtitle {23rd European Symposium on Artificial Neural Networks, Computational Intelligence and Machine Learning Bruges} {23rd european symposium on artificial neural networks, computational intelligence and machine learning bruges}\ (\BPG~94).
\PrintBackRefs{\CurrentBib}

\bibitem [\protect \citeauthoryear {%
Matteson%
\ \BBA {} James%
}{%
Matteson%
\ \BBA {} James%
}{%
{\protect \APACyear {2014}}%
}]{%
Matteson2014AData}
\APACinsertmetastar {%
Matteson2014AData}%
\begin{APACrefauthors}%
Matteson, D.S.%
\BCBT {}\ \BBA {} James, N.A.%
\end{APACrefauthors}%
\unskip\
\newblock
\APACrefYearMonthDay{2014}{}{}.
\newblock
{\BBOQ}\APACrefatitle {{A nonparametric approach for multiple change point analysis of multivariate data}} {{A nonparametric approach for multiple change point analysis of multivariate data}}.{\BBCQ}
\newblock
\APACjournalVolNumPages{Journal of the American Statistical Association}{109}{505}{334--345,}
\newblock
\begin{APACrefDOI} \doi{10.1080/01621459.2013.849605} \end{APACrefDOI}
\newblock

\newblock

\PrintBackRefs{\CurrentBib}

\bibitem [\protect \citeauthoryear {%
Maudsley%
\ \protect \BOthers {.}}{%
Maudsley%
\ \protect \BOthers {.}}{%
{\protect \APACyear {2021}}%
}]{%
Maudsley2021AdvancedRecommendations}
\APACinsertmetastar {%
Maudsley2021AdvancedRecommendations}%
\begin{APACrefauthors}%
Maudsley, A.A.%
, Andronesi, O.C.%
, Barker, P.B.%
, Bizzi, A.%
, Bogner, W.%
, Henning, A.%
\BDBL {}Soher, B.J.%
\end{APACrefauthors}%
\unskip\
\newblock
\APACrefYearMonthDay{2021}{5}{}.
\newblock
{\BBOQ}\APACrefatitle {{Advanced magnetic resonance spectroscopic neuroimaging: Experts' consensus recommendations}} {{Advanced magnetic resonance spectroscopic neuroimaging: Experts' consensus recommendations}}.{\BBCQ}
\newblock
\APACjournalVolNumPages{NMR in Biomedicine}{34}{5}{,}
\newblock
\begin{APACrefDOI} \doi{10.1002/nbm.4309} \end{APACrefDOI}
\newblock

\newblock

\PrintBackRefs{\CurrentBib}

\bibitem [\protect \citeauthoryear {%
Maudsley%
\ \protect \BOthers {.}}{%
Maudsley%
\ \protect \BOthers {.}}{%
{\protect \APACyear {2006}}%
}]{%
Maudsley2006ComprehensiveImaging}
\APACinsertmetastar {%
Maudsley2006ComprehensiveImaging}%
\begin{APACrefauthors}%
Maudsley, A.A.%
, Darkazanli, A.%
, Alger, J.R.%
, Hall, L.O.%
, Schuff, N.%
, Studholme, C.%
\BDBL {}Zhu, X.%
\end{APACrefauthors}%
\unskip\
\newblock
\APACrefYearMonthDay{2006}{6}{}.
\newblock
{\BBOQ}\APACrefatitle {{Comprehensive processing, display and analysis for in vivo MR spectroscopic imaging}} {{Comprehensive processing, display and analysis for in vivo MR spectroscopic imaging}}.{\BBCQ}
\newblock
\APACjournalVolNumPages{NMR in Biomedicine}{19}{4}{492--503,}
\newblock
\begin{APACrefDOI} \doi{10.1002/nbm.1025} \end{APACrefDOI}
\newblock

\newblock

\PrintBackRefs{\CurrentBib}

\bibitem [\protect \citeauthoryear {%
McGonigle%
\ \BBA {} Cho%
}{%
McGonigle%
\ \BBA {} Cho%
}{%
{\protect \APACyear {2025}}%
}]{%
McGonigle2025NonparametricFunctions}
\APACinsertmetastar {%
McGonigle2025NonparametricFunctions}%
\begin{APACrefauthors}%
McGonigle, E.T.%
\BCBT {}\ \BBA {} Cho, H.%
\end{APACrefauthors}%
\unskip\
\newblock
\APACrefYearMonthDay{2025}{8}{}.
\newblock
{\BBOQ}\APACrefatitle {{Nonparametric data segmentation in multivariate time series via joint characteristic functions}} {{Nonparametric data segmentation in multivariate time series via joint characteristic functions}}.{\BBCQ}
\newblock
\APACjournalVolNumPages{arXiv}{}{}{,}
\newblock
\begin{APACrefDOI} \doi{10.1093/biomet/asaf024} \end{APACrefDOI}
\newblock

\newblock

\PrintBackRefs{\CurrentBib}

\bibitem [\protect \citeauthoryear {%
Nam%
\ \protect \BOthers {.}}{%
Nam%
\ \protect \BOthers {.}}{%
{\protect \APACyear {2023}}%
}]{%
Nam2023DeuteriumT}
\APACinsertmetastar {%
Nam2023DeuteriumT}%
\begin{APACrefauthors}%
Nam, K.M.%
, Gursan, A.%
, Bhogal, A.A.%
, Wijnen, J.P.%
, Klomp, D.W.J.%
, Prompers, J.J.%
\BCBL {} Hendriks, A.D.%
\end{APACrefauthors}%
\unskip\
\newblock
\APACrefYearMonthDay{2023}{9}{}.
\newblock
{\BBOQ}\APACrefatitle {{Deuterium echo‐planar spectroscopic imaging (EPSI) in the human liver in vivo at 7 T}} {{Deuterium echo‐planar spectroscopic imaging (EPSI) in the human liver in vivo at 7 T}}.{\BBCQ}
\newblock
\APACjournalVolNumPages{Magnetic Resonance in Medicine}{90}{3}{863--874,}
\newblock
\begin{APACrefDOI} \doi{10.1002/mrm.29696} \end{APACrefDOI}
\newblock

\newblock

\PrintBackRefs{\CurrentBib}

\bibitem [\protect \citeauthoryear {%
Near%
\ \protect \BOthers {.}}{%
Near%
\ \protect \BOthers {.}}{%
{\protect \APACyear {2021}}%
}]{%
Near2021PreprocessingRecommendations}
\APACinsertmetastar {%
Near2021PreprocessingRecommendations}%
\begin{APACrefauthors}%
Near, J.%
, Harris, A.D.%
, Juchem, C.%
, Kreis, R.%
, Marja{\'{n}}ska, M.%
, {\"{O}}z, G.%
\BDBL {}Gasparovic, C.%
\end{APACrefauthors}%
\unskip\
\newblock
\APACrefYearMonthDay{2021}{5}{}.
\newblock
{\BBOQ}\APACrefatitle {{Preprocessing, analysis and quantification in single‐voxel magnetic resonance spectroscopy: experts' consensus recommendations}} {{Preprocessing, analysis and quantification in single‐voxel magnetic resonance spectroscopy: experts' consensus recommendations}}.{\BBCQ}
\newblock
\APACjournalVolNumPages{NMR in Biomedicine}{34}{5}{,}
\newblock
\begin{APACrefDOI} \doi{10.1002/nbm.4257} \end{APACrefDOI}
\newblock

\newblock

\PrintBackRefs{\CurrentBib}

\bibitem [\protect \citeauthoryear {%
Nguyen%
\ \BBA {} Hocking%
}{%
Nguyen%
\ \BBA {} Hocking%
}{%
{\protect \APACyear {2025}}%
}]{%
Nguyen2025PenaltyPerceptron}
\APACinsertmetastar {%
Nguyen2025PenaltyPerceptron}%
\begin{APACrefauthors}%
Nguyen, T.L.%
\BCBT {}\ \BBA {} Hocking, T.D.%
\end{APACrefauthors}%
\unskip\
\newblock
\APACrefYearMonthDay{2025}{6}{}.
\newblock
{\BBOQ}\APACrefatitle {{Penalty Learning for Optimal Partitioning using Multilayer Perceptron}} {{Penalty Learning for Optimal Partitioning using Multilayer Perceptron}}.{\BBCQ}
\newblock
\APACjournalVolNumPages{arXiv}{}{}{,}
\newblock
\begin{APACrefDOI} \doi{10.1007/s11222-025-10680-0} \end{APACrefDOI}
\newblock

\newblock

\PrintBackRefs{\CurrentBib}

\bibitem [\protect \citeauthoryear {%
Nix%
\ \BBA {} Weigend%
}{%
Nix%
\ \BBA {} Weigend%
}{%
{\protect \APACyear {1994}}%
}]{%
Nix1994EstimatingDistribution}
\APACinsertmetastar {%
Nix1994EstimatingDistribution}%
\begin{APACrefauthors}%
Nix, D.%
\BCBT {}\ \BBA {} Weigend, A.%
\end{APACrefauthors}%
\unskip\
\newblock
\APACrefYearMonthDay{1994}{}{}.
\newblock
{\BBOQ}\APACrefatitle {{Estimating the mean and variance of the target probability distribution}} {{Estimating the mean and variance of the target probability distribution}}.{\BBCQ}
\newblock
 \APACrefbtitle {Proceedings of 1994 IEEE International Conference on Neural Networks (ICNN'94)} {Proceedings of 1994 ieee international conference on neural networks (icnn'94)}\ (\BVOL~1, \BPGS\ 55--60).
\newblock
\APACaddressPublisher{}{IEEE}.
\PrintBackRefs{\CurrentBib}

\bibitem [\protect \citeauthoryear {%
Obst%
, Wang%
\BCBL {}\ \BBA {} Prokopenko%
}{%
Obst%
\ \protect \BOthers {.}}{%
{\protect \APACyear {2008}}%
}]{%
Obst2008UsingMines}
\APACinsertmetastar {%
Obst2008UsingMines}%
\begin{APACrefauthors}%
Obst, O.%
, Wang, X.R.%
\BCBL {} Prokopenko, M.%
\end{APACrefauthors}%
\unskip\
\newblock
\APACrefYearMonthDay{2008}{}{}.
\newblock
{\BBOQ}\APACrefatitle {{Using echo state networks for anomaly detection in underground coal mines}} {{Using echo state networks for anomaly detection in underground coal mines}}.{\BBCQ}
\newblock
 \APACrefbtitle {Proceedings - 2008 International Conference on Information Processing in Sensor Networks, IPSN 2008} {Proceedings - 2008 international conference on information processing in sensor networks, ipsn 2008}\ (\BPGS\ 219--229).
\PrintBackRefs{\CurrentBib}

\bibitem [\protect \citeauthoryear {%
Oord%
, Li%
\BCBL {}\ \BBA {} Vinyals%
}{%
Oord%
\ \protect \BOthers {.}}{%
{\protect \APACyear {2018}}%
}]{%
VanDenOord2018}
\APACinsertmetastar {%
VanDenOord2018}%
\begin{APACrefauthors}%
Oord, A.v.d.%
, Li, Y.%
\BCBL {} Vinyals, O.%
\end{APACrefauthors}%
\unskip\
\newblock
\APACrefYearMonthDay{2018}{7}{}.
\newblock
{\BBOQ}\APACrefatitle {{Representation Learning with Contrastive Predictive Coding}} {{Representation Learning with Contrastive Predictive Coding}}.{\BBCQ}
\newblock
\APACjournalVolNumPages{arXiv}{}{}{,}
\newblock

\newblock

\PrintBackRefs{\CurrentBib}

\bibitem [\protect \citeauthoryear {%
{\"{O}}z%
\ \protect \BOthers {.}}{%
{\"{O}}z%
\ \protect \BOthers {.}}{%
{\protect \APACyear {2014}}%
}]{%
Oz2014ClinicalDisorders}
\APACinsertmetastar {%
Oz2014ClinicalDisorders}%
\begin{APACrefauthors}%
{\"{O}}z, G.%
, Alger, J.R.%
, Barker, P.B.%
, Bartha, R.%
, Bizzi, A.%
, Boesch, C.%
\BDBL {}Kauppinen, R.A.%
\end{APACrefauthors}%
\unskip\
\newblock
\APACrefYearMonthDay{2014}{3}{}.
\newblock
{\BBOQ}\APACrefatitle {{Clinical Proton MR Spectroscopy in Central Nervous System Disorders}} {{Clinical Proton MR Spectroscopy in Central Nervous System Disorders}}.{\BBCQ}
\newblock
\APACjournalVolNumPages{Radiology}{270}{3}{658--679,}
\newblock
\begin{APACrefDOI} \doi{10.1148/radiol.13130531} \end{APACrefDOI}
\newblock

\newblock

\PrintBackRefs{\CurrentBib}

\bibitem [\protect \citeauthoryear {%
Page%
}{%
Page%
}{%
{\protect \APACyear {1954}}%
}]{%
Page1954ContinuousSchemes}
\APACinsertmetastar {%
Page1954ContinuousSchemes}%
\begin{APACrefauthors}%
Page, E.S.%
\end{APACrefauthors}%
\unskip\
\newblock
\APACrefYearMonthDay{1954}{6}{}.
\newblock
{\BBOQ}\APACrefatitle {{Continuous Inspection Schemes}} {{Continuous Inspection Schemes}}.{\BBCQ}
\newblock
\APACjournalVolNumPages{Biometrika}{41}{1/2}{100,}
\newblock
\begin{APACrefDOI} \doi{10.2307/2333009} \end{APACrefDOI}
\newblock

\newblock

\PrintBackRefs{\CurrentBib}

\bibitem [\protect \citeauthoryear {%
Pang%
, Shen%
, Cao%
\BCBL {}\ \BBA {} Hengel%
}{%
Pang%
\ \protect \BOthers {.}}{%
{\protect \APACyear {2022}}%
}]{%
Pang2022DeepDetection}
\APACinsertmetastar {%
Pang2022DeepDetection}%
\begin{APACrefauthors}%
Pang, G.%
, Shen, C.%
, Cao, L.%
\BCBL {} Hengel, A.V.D.%
\end{APACrefauthors}%
\unskip\
\newblock
\APACrefYearMonthDay{2022}{3}{}.
\newblock
{\BBOQ}\APACrefatitle {{Deep Learning for Anomaly Detection}} {{Deep Learning for Anomaly Detection}}.{\BBCQ}
\newblock
\APACjournalVolNumPages{ACM Computing Surveys}{54}{2}{1--38,}
\newblock
\begin{APACrefDOI} \doi{10.1145/3439950} \end{APACrefDOI}
\newblock

\newblock

\PrintBackRefs{\CurrentBib}

\bibitem [\protect \citeauthoryear {%
Pang%
, Ting%
, Albrecht%
\BCBL {}\ \BBA {} Jin%
}{%
Pang%
\ \protect \BOthers {.}}{%
{\protect \APACyear {2016}}%
}]{%
Pang2016ZERO++:Sets}
\APACinsertmetastar {%
Pang2016ZERO++:Sets}%
\begin{APACrefauthors}%
Pang, G.%
, Ting, K.M.%
, Albrecht, D.%
\BCBL {} Jin, H.%
\end{APACrefauthors}%
\unskip\
\newblock
\APACrefYearMonthDay{2016}{12}{}.
\newblock
{\BBOQ}\APACrefatitle {{ZERO++: Harnessing the Power of Zero Appearances to Detect Anomalies in Large-Scale Data Sets}} {{ZERO++: Harnessing the Power of Zero Appearances to Detect Anomalies in Large-Scale Data Sets}}.{\BBCQ}
\newblock
\APACjournalVolNumPages{Journal of Artificial Intelligence Research}{57}{}{593--620,}
\newblock
\begin{APACrefDOI} \doi{10.1613/jair.5228} \end{APACrefDOI}
\newblock

\newblock

\PrintBackRefs{\CurrentBib}

\bibitem [\protect \citeauthoryear {%
Paparrizos%
, Boniol%
, Liu%
\BCBL {}\ \BBA {} Palpanas%
}{%
Paparrizos%
\ \protect \BOthers {.}}{%
{\protect \APACyear {2025}}%
}]{%
Paparrizos2025AdvancesMeasures}
\APACinsertmetastar {%
Paparrizos2025AdvancesMeasures}%
\begin{APACrefauthors}%
Paparrizos, J.%
, Boniol, P.%
, Liu, Q.%
\BCBL {} Palpanas, T.%
\end{APACrefauthors}%
\unskip\
\newblock
\APACrefYearMonthDay{2025}{8}{}.
\newblock
{\BBOQ}\APACrefatitle {{Advances in Time-Series Anomaly Detection: Algorithms, Benchmarks, and Evaluation Measures}} {{Advances in Time-Series Anomaly Detection: Algorithms, Benchmarks, and Evaluation Measures}}.{\BBCQ}
\newblock
 \APACrefbtitle {Proceedings of the ACM SIGKDD International Conference on Knowledge Discovery and Data Mining} {Proceedings of the acm sigkdd international conference on knowledge discovery and data mining}\ (\BVOL~2, \BPGS\ 6151--6161).
\newblock
\APACaddressPublisher{}{Association for Computing Machinery}.
\PrintBackRefs{\CurrentBib}

\bibitem [\protect \citeauthoryear {%
Pedrosa~de Barros%
, McKinley%
, Wiest%
\BCBL {}\ \BBA {} Slotboom%
}{%
Pedrosa~de Barros%
\ \protect \BOthers {.}}{%
{\protect \APACyear {2017}}%
}]{%
PedrosadeBarros2017ImprovingData}
\APACinsertmetastar {%
PedrosadeBarros2017ImprovingData}%
\begin{APACrefauthors}%
Pedrosa~de Barros, N.%
, McKinley, R.%
, Wiest, R.%
\BCBL {} Slotboom, J.%
\end{APACrefauthors}%
\unskip\
\newblock
\APACrefYearMonthDay{2017}{12}{}.
\newblock
{\BBOQ}\APACrefatitle {{Improving labeling efficiency in automatic quality control of MRSI data}} {{Improving labeling efficiency in automatic quality control of MRSI data}}.{\BBCQ}
\newblock
\APACjournalVolNumPages{Magnetic Resonance in Medicine}{78}{6}{2399--2405,}
\newblock
\begin{APACrefDOI} \doi{10.1002/mrm.26618} \end{APACrefDOI}
\newblock

\newblock

\PrintBackRefs{\CurrentBib}

\bibitem [\protect \citeauthoryear {%
Pelletier%
\ \protect \BOthers {.}}{%
Pelletier%
\ \protect \BOthers {.}}{%
{\protect \APACyear {2002}}%
}]{%
Pelletier20023-DBrain}
\APACinsertmetastar {%
Pelletier20023-DBrain}%
\begin{APACrefauthors}%
Pelletier, D.%
, Nelson, S.%
, Grenier, D.%
, Lu, Y.%
, Genain, C.%
\BCBL {} Goodkin, D.%
\end{APACrefauthors}%
\unskip\
\newblock
\APACrefYearMonthDay{2002}{10}{}.
\newblock
{\BBOQ}\APACrefatitle {{3-D echo planar 1HMRS imaging in MS: metabolite comparison from supratentorial vs. central brain}} {{3-D echo planar 1HMRS imaging in MS: metabolite comparison from supratentorial vs. central brain}}.{\BBCQ}
\newblock
\APACjournalVolNumPages{Magnetic Resonance Imaging}{20}{8}{599--606,}
\newblock
\begin{APACrefDOI} \doi{10.1016/S0730-725X(02)00533-7} \end{APACrefDOI}
\newblock

\newblock

\PrintBackRefs{\CurrentBib}

\bibitem [\protect \citeauthoryear {%
Posse%
, Otazo%
, Dager%
\BCBL {}\ \BBA {} Alger%
}{%
Posse%
\ \protect \BOthers {.}}{%
{\protect \APACyear {2013}}%
}]{%
Posse2013MRAdvances}
\APACinsertmetastar {%
Posse2013MRAdvances}%
\begin{APACrefauthors}%
Posse, S.%
, Otazo, R.%
, Dager, S.R.%
\BCBL {} Alger, J.%
\end{APACrefauthors}%
\unskip\
\newblock
\APACrefYearMonthDay{2013}{6}{}.
\newblock
{\BBOQ}\APACrefatitle {{MR spectroscopic imaging: Principles and recent advances}} {{MR spectroscopic imaging: Principles and recent advances}}.{\BBCQ}
\newblock
\APACjournalVolNumPages{Journal of Magnetic Resonance Imaging}{37}{6}{1301--1325,}
\newblock
\begin{APACrefDOI} \doi{10.1002/jmri.23945} \end{APACrefDOI}
\newblock

\newblock

\PrintBackRefs{\CurrentBib}

\bibitem [\protect \citeauthoryear {%
Provencher%
}{%
Provencher%
}{%
{\protect \APACyear {1993}}%
}]{%
Provencher1993EstimationSpectra}
\APACinsertmetastar {%
Provencher1993EstimationSpectra}%
\begin{APACrefauthors}%
Provencher, S.W.%
\end{APACrefauthors}%
\unskip\
\newblock
\APACrefYearMonthDay{1993}{12}{}.
\newblock
{\BBOQ}\APACrefatitle {{Estimation of metabolite concentrations from localized in vivo proton NMR spectra}} {{Estimation of metabolite concentrations from localized in vivo proton NMR spectra}}.{\BBCQ}
\newblock
\APACjournalVolNumPages{Magnetic Resonance in Medicine}{30}{6}{672--679,}
\newblock
\begin{APACrefDOI} \doi{10.1002/mrm.1910300604} \end{APACrefDOI}
\newblock

\newblock

\PrintBackRefs{\CurrentBib}

\bibitem [\protect \citeauthoryear {%
Ryzhikov%
, Hushchyn%
\BCBL {}\ \BBA {} Derkach%
}{%
Ryzhikov%
\ \protect \BOthers {.}}{%
{\protect \APACyear {2023}}%
}]{%
Ryzhikov2023LatentDetection}
\APACinsertmetastar {%
Ryzhikov2023LatentDetection}%
\begin{APACrefauthors}%
Ryzhikov, A.%
, Hushchyn, M.%
\BCBL {} Derkach, D.%
\end{APACrefauthors}%
\unskip\
\newblock
\APACrefYearMonthDay{2023}{}{}.
\newblock
{\BBOQ}\APACrefatitle {{Latent Stochastic Differential Equations for Change Point Detection}} {{Latent Stochastic Differential Equations for Change Point Detection}}.{\BBCQ}
\newblock
\APACjournalVolNumPages{IEEE Access}{11}{}{104700--104711,}
\newblock
\begin{APACrefDOI} \doi{10.1109/ACCESS.2023.3318318} \end{APACrefDOI}
\newblock

\newblock

\PrintBackRefs{\CurrentBib}

\bibitem [\protect \citeauthoryear {%
Schlegl%
, Seeb{\"{o}}ck%
, Waldstein%
, Langs%
\BCBL {}\ \BBA {} Schmidt-Erfurth%
}{%
Schlegl%
\ \protect \BOthers {.}}{%
{\protect \APACyear {2019}}%
}]{%
Schlegl2019F-AnoGAN:Networks}
\APACinsertmetastar {%
Schlegl2019F-AnoGAN:Networks}%
\begin{APACrefauthors}%
Schlegl, T.%
, Seeb{\"{o}}ck, P.%
, Waldstein, S.M.%
, Langs, G.%
\BCBL {} Schmidt-Erfurth, U.%
\end{APACrefauthors}%
\unskip\
\newblock
\APACrefYearMonthDay{2019}{5}{}.
\newblock
{\BBOQ}\APACrefatitle {{f-AnoGAN: Fast unsupervised anomaly detection with generative adversarial networks}} {{f-AnoGAN: Fast unsupervised anomaly detection with generative adversarial networks}}.{\BBCQ}
\newblock
\APACjournalVolNumPages{Medical Image Analysis}{54}{}{30--44,}
\newblock
\begin{APACrefDOI} \doi{10.1016/j.media.2019.01.010} \end{APACrefDOI}
\newblock

\newblock

\PrintBackRefs{\CurrentBib}

\bibitem [\protect \citeauthoryear {%
Simonyan%
\ \BBA {} Zisserman%
}{%
Simonyan%
\ \BBA {} Zisserman%
}{%
{\protect \APACyear {2014}}%
}]{%
Simonyan2014VeryRecognition}
\APACinsertmetastar {%
Simonyan2014VeryRecognition}%
\begin{APACrefauthors}%
Simonyan, K.%
\BCBT {}\ \BBA {} Zisserman, A.%
\end{APACrefauthors}%
\unskip\
\newblock
\APACrefYearMonthDay{2014}{9}{}.
\newblock
{\BBOQ}\APACrefatitle {{Very Deep Convolutional Networks for Large-Scale Image Recognition}} {{Very Deep Convolutional Networks for Large-Scale Image Recognition}}.{\BBCQ}
\newblock
\APACjournalVolNumPages{ICLR}{}{}{,}
\newblock

\newblock

\PrintBackRefs{\CurrentBib}

\bibitem [\protect \citeauthoryear {%
Sluijterman%
, Cator%
\BCBL {}\ \BBA {} Heskes%
}{%
Sluijterman%
\ \protect \BOthers {.}}{%
{\protect \APACyear {2023}}%
}]{%
Sluijterman2023OptimalNetworks}
\APACinsertmetastar {%
Sluijterman2023OptimalNetworks}%
\begin{APACrefauthors}%
Sluijterman, L.%
, Cator, E.%
\BCBL {} Heskes, T.%
\end{APACrefauthors}%
\unskip\
\newblock
\APACrefYearMonthDay{2023}{2}{}.
\newblock
{\BBOQ}\APACrefatitle {{Optimal Training of Mean Variance Estimation Neural Networks}} {{Optimal Training of Mean Variance Estimation Neural Networks}}.{\BBCQ}
\newblock
\APACjournalVolNumPages{arXiv}{}{}{,}
\newblock

\newblock

\PrintBackRefs{\CurrentBib}

\bibitem [\protect \citeauthoryear {%
Snyder%
\ \BBA {} Withers%
}{%
Snyder%
\ \BBA {} Withers%
}{%
{\protect \APACyear {1983}}%
}]{%
Snyder1983ExponentialCorrection}
\APACinsertmetastar {%
Snyder1983ExponentialCorrection}%
\begin{APACrefauthors}%
Snyder, R.D.%
\BCBT {}\ \BBA {} Withers, S.J.%
\end{APACrefauthors}%
\unskip\
\newblock
\APACrefYear{1983}.
\newblock
\APACrefbtitle {{Exponential smoothing with finite sample correction}} {{Exponential smoothing with finite sample correction}}.
\newblock
\APACaddressPublisher{}{Dept. of Econometrics and Operations Research, Faculty of Economics and Politics, Monash University}.
\PrintBackRefs{\CurrentBib}

\bibitem [\protect \citeauthoryear {%
Su%
\ \protect \BOthers {.}}{%
Su%
\ \protect \BOthers {.}}{%
{\protect \APACyear {2019}}%
}]{%
Su2019RobustNetwork}
\APACinsertmetastar {%
Su2019RobustNetwork}%
\begin{APACrefauthors}%
Su, Y.%
, Liu, R.%
, Zhao, Y.%
, Sun, W.%
, Niu, C.%
\BCBL {} Pei, D.%
\end{APACrefauthors}%
\unskip\
\newblock
\APACrefYearMonthDay{2019}{7}{}.
\newblock
{\BBOQ}\APACrefatitle {{Robust anomaly detection for multivariate time series through stochastic recurrent neural network}} {{Robust anomaly detection for multivariate time series through stochastic recurrent neural network}}.{\BBCQ}
\newblock
 \APACrefbtitle {Proceedings of the ACM SIGKDD International Conference on Knowledge Discovery and Data Mining} {Proceedings of the acm sigkdd international conference on knowledge discovery and data mining}\ (\BPGS\ 2828--2837).
\newblock
\APACaddressPublisher{}{Association for Computing Machinery}.
\PrintBackRefs{\CurrentBib}

\bibitem [\protect \citeauthoryear {%
Szegedy%
, Vanhoucke%
, Ioffe%
, Shlens%
\BCBL {}\ \BBA {} Wojna%
}{%
Szegedy%
\ \protect \BOthers {.}}{%
{\protect \APACyear {2016}}%
}]{%
Szegedy2016RethinkingVision}
\APACinsertmetastar {%
Szegedy2016RethinkingVision}%
\begin{APACrefauthors}%
Szegedy, C.%
, Vanhoucke, V.%
, Ioffe, S.%
, Shlens, J.%
\BCBL {} Wojna, Z.%
\end{APACrefauthors}%
\unskip\
\newblock
\APACrefYearMonthDay{2016}{6}{}.
\newblock
{\BBOQ}\APACrefatitle {{Rethinking the Inception Architecture for Computer Vision}} {{Rethinking the Inception Architecture for Computer Vision}}.{\BBCQ}
\newblock
 \APACrefbtitle {2016 IEEE Conference on Computer Vision and Pattern Recognition (CVPR)} {2016 ieee conference on computer vision and pattern recognition (cvpr)}\ (\BPGS\ 2818--2826).
\newblock
\APACaddressPublisher{}{IEEE}.
\PrintBackRefs{\CurrentBib}

\bibitem [\protect \citeauthoryear {%
Takeuchi%
\ \BBA {} Yamanishi%
}{%
Takeuchi%
\ \BBA {} Yamanishi%
}{%
{\protect \APACyear {2006}}%
}]{%
Takeuchi2006ASeries}
\APACinsertmetastar {%
Takeuchi2006ASeries}%
\begin{APACrefauthors}%
Takeuchi, J.%
\BCBT {}\ \BBA {} Yamanishi, K.%
\end{APACrefauthors}%
\unskip\
\newblock
\APACrefYearMonthDay{2006}{4}{}.
\newblock
{\BBOQ}\APACrefatitle {{A unifying framework for detecting outliers and change points from time series}} {{A unifying framework for detecting outliers and change points from time series}}.{\BBCQ}
\newblock
\APACjournalVolNumPages{IEEE Transactions on Knowledge and Data Engineering}{18}{4}{482--492,}
\newblock
\begin{APACrefDOI} \doi{10.1109/TKDE.2006.1599387} \end{APACrefDOI}
\newblock

\newblock

\PrintBackRefs{\CurrentBib}

\bibitem [\protect \citeauthoryear {%
Theiler%
\ \BBA {} Perkins%
}{%
Theiler%
\ \BBA {} Perkins%
}{%
{\protect \APACyear {2006}}%
}]{%
Theiler2006ProposedDetection}
\APACinsertmetastar {%
Theiler2006ProposedDetection}%
\begin{APACrefauthors}%
Theiler, J.%
\BCBT {}\ \BBA {} Perkins, S.%
\end{APACrefauthors}%
\unskip\
\newblock
\APACrefYearMonthDay{2006}{}{}.
\newblock
{\BBOQ}\APACrefatitle {{Proposed framework for anomalous change detection}} {{Proposed framework for anomalous change detection}}.{\BBCQ}
\newblock
 \APACrefbtitle {ICML Workshop on Machine Learning Algorithms for Surveillance and Event Detection} {Icml workshop on machine learning algorithms for surveillance and event detection}\ (\BPGS\ 7--14).
\PrintBackRefs{\CurrentBib}

\bibitem [\protect \citeauthoryear {%
Truong%
, Oudre%
\BCBL {}\ \BBA {} Vayatis%
}{%
Truong%
\ \protect \BOthers {.}}{%
{\protect \APACyear {2020}}%
}]{%
Truong2020SelectiveMethods}
\APACinsertmetastar {%
Truong2020SelectiveMethods}%
\begin{APACrefauthors}%
Truong, C.%
, Oudre, L.%
\BCBL {} Vayatis, N.%
\end{APACrefauthors}%
\unskip\
\newblock
\APACrefYearMonthDay{2020}{2}{}.
\newblock
{\BBOQ}\APACrefatitle {{Selective review of offline change point detection methods}} {{Selective review of offline change point detection methods}}.{\BBCQ}
\newblock
\APACjournalVolNumPages{Signal Processing}{167}{}{107299,}
\newblock
\begin{APACrefDOI} \doi{10.1016/j.sigpro.2019.107299} \end{APACrefDOI}
\newblock

\newblock

\PrintBackRefs{\CurrentBib}

\bibitem [\protect \citeauthoryear {%
Tuli%
, Casale%
\BCBL {}\ \BBA {} Jennings%
}{%
Tuli%
\ \protect \BOthers {.}}{%
{\protect \APACyear {2022}}%
}]{%
Tuli2022}
\APACinsertmetastar {%
Tuli2022}%
\begin{APACrefauthors}%
Tuli, S.%
, Casale, G.%
\BCBL {} Jennings, N.R.%
\end{APACrefauthors}%
\unskip\
\newblock
\APACrefYearMonthDay{2022}{5}{}.
\newblock
{\BBOQ}\APACrefatitle {{TranAD: Deep Transformer Networks for Anomaly Detection in Multivariate Time Series Data}} {{TranAD: Deep Transformer Networks for Anomaly Detection in Multivariate Time Series Data}}.{\BBCQ}
\newblock
\APACjournalVolNumPages{arXiv}{}{}{,}
\newblock

\newblock

\PrintBackRefs{\CurrentBib}

\bibitem [\protect \citeauthoryear {%
van~de Sande%
\ \protect \BOthers {.}}{%
van~de Sande%
\ \protect \BOthers {.}}{%
{\protect \APACyear {2023}}%
}]{%
vandeSande2023AWorkflow}
\APACinsertmetastar {%
vandeSande2023AWorkflow}%
\begin{APACrefauthors}%
van~de Sande, D.M.J.%
, Merkofer, J.P.%
, Amirrajab, S.%
, Veta, M.%
, van Sloun, R.J.G.%
, Versluis, M.J.%
\BDBL {}Breeuwer, M.%
\end{APACrefauthors}%
\unskip\
\newblock
\APACrefYearMonthDay{2023}{10}{}.
\newblock
{\BBOQ}\APACrefatitle {{A review of machine learning applications for the proton MR spectroscopy workflow}} {{A review of machine learning applications for the proton MR spectroscopy workflow}}.{\BBCQ}
\newblock
\APACjournalVolNumPages{Magnetic Resonance in Medicine}{90}{4}{1253--1270,}
\newblock
\begin{APACrefDOI} \doi{10.1002/mrm.29793} \end{APACrefDOI}
\newblock

\newblock

\PrintBackRefs{\CurrentBib}

\bibitem [\protect \citeauthoryear {%
Wald%
}{%
Wald%
}{%
{\protect \APACyear {1945}}%
}]{%
Wald1945SequentialHypotheses}
\APACinsertmetastar {%
Wald1945SequentialHypotheses}%
\begin{APACrefauthors}%
Wald, A.%
\end{APACrefauthors}%
\unskip\
\newblock
\APACrefYearMonthDay{1945}{6}{}.
\newblock
{\BBOQ}\APACrefatitle {{Sequential Tests of Statistical Hypotheses}} {{Sequential Tests of Statistical Hypotheses}}.{\BBCQ}
\newblock
\APACjournalVolNumPages{The Annals of Mathematical Statistics}{16}{2}{117--186,}
\newblock
\begin{APACrefDOI} \doi{10.1214/aoms/1177731118} \end{APACrefDOI}
\newblock
\begin{APACrefURL} {http://projecteuclid.org/euclid.aoms/1177731118} \end{APACrefURL}
\newblock

\newblock

\PrintBackRefs{\CurrentBib}

\bibitem [\protect \citeauthoryear {%
Wang%
, Lin%
, Mishra%
\BCBL {}\ \BBA {} Sriharsha%
}{%
Wang%
\ \protect \BOthers {.}}{%
{\protect \APACyear {2021}}%
}]{%
Wang2021OnlineBudget}
\APACinsertmetastar {%
Wang2021OnlineBudget}%
\begin{APACrefauthors}%
Wang, Z.%
, Lin, X.%
, Mishra, A.%
\BCBL {} Sriharsha, R.%
\end{APACrefauthors}%
\unskip\
\newblock
\APACrefYearMonthDay{2021}{12}{}.
\newblock
{\BBOQ}\APACrefatitle {{Online Changepoint Detection on a Budget}} {{Online Changepoint Detection on a Budget}}.{\BBCQ}
\newblock
 \APACrefbtitle {2021 International Conference on Data Mining Workshops (ICDMW)} {2021 international conference on data mining workshops (icdmw)}\ (\BVOL\ 2021-December, \BPGS\ 414--420).
\newblock
\APACaddressPublisher{}{IEEE}.
\PrintBackRefs{\CurrentBib}

\bibitem [\protect \citeauthoryear {%
D.~Wu%
, Gundimeda%
, Mou%
\BCBL {}\ \BBA {} Quinn%
}{%
D.~Wu%
\ \protect \BOthers {.}}{%
{\protect \APACyear {2024}}%
}]{%
Wu2024UnsupervisedSeries}
\APACinsertmetastar {%
Wu2024UnsupervisedSeries}%
\begin{APACrefauthors}%
Wu, D.%
, Gundimeda, S.%
, Mou, S.%
\BCBL {} Quinn, C.J.%
\end{APACrefauthors}%
\unskip\
\newblock
\APACrefYearMonthDay{2024}{}{}.
\newblock
{\BBOQ}\APACrefatitle {{Unsupervised Change Point Detection in Multivariate Time Series}} {{Unsupervised Change Point Detection in Multivariate Time Series}}.{\BBCQ}
\newblock
 \APACrefbtitle {Proceedings of The 27th International Conference on Artificial Intelligence and Statistics} {Proceedings of the 27th international conference on artificial intelligence and statistics}\ (\BPGS\ 3844--3852).
\newblock
\APACaddressPublisher{}{PMLR}.
\PrintBackRefs{\CurrentBib}

\bibitem [\protect \citeauthoryear {%
H.~Wu%
, Schafer%
\BCBL {}\ \BBA {} Matteson%
}{%
H.~Wu%
\ \protect \BOthers {.}}{%
{\protect \APACyear {2024}}%
}]{%
Wu2024TrendScoring}
\APACinsertmetastar {%
Wu2024TrendScoring}%
\begin{APACrefauthors}%
Wu, H.%
, Schafer, T.L.J.%
\BCBL {} Matteson, D.S.%
\end{APACrefauthors}%
\unskip\
\newblock
\APACrefYearMonthDay{2024}{3}{}.
\newblock
{\BBOQ}\APACrefatitle {{Trend and Variance Adaptive Bayesian Changepoint Analysis {\&} Local Outlier Scoring}} {{Trend and Variance Adaptive Bayesian Changepoint Analysis {\&} Local Outlier Scoring}}.{\BBCQ}
\newblock
\APACjournalVolNumPages{arXiv}{}{}{,}
\newblock

\newblock

\PrintBackRefs{\CurrentBib}

\bibitem [\protect \citeauthoryear {%
Y.~Wu%
\ \protect \BOthers {.}}{%
Y.~Wu%
\ \protect \BOthers {.}}{%
{\protect \APACyear {2023}}%
}]{%
Wu2023CLformer:Structures}
\APACinsertmetastar {%
Wu2023CLformer:Structures}%
\begin{APACrefauthors}%
Wu, Y.%
, Dong, Y.%
, Zhu, W.%
, Zhang, J.%
, Liu, S.%
, Lu, D.%
\BDBL {}Li, Y.%
\end{APACrefauthors}%
\unskip\
\newblock
\APACrefYearMonthDay{2023}{11}{}.
\newblock
{\BBOQ}\APACrefatitle {{CLformer: Constraint-based Locality enhanced Transformer for anomaly detection of ancient building structures}} {{CLformer: Constraint-based Locality enhanced Transformer for anomaly detection of ancient building structures}}.{\BBCQ}
\newblock
\APACjournalVolNumPages{Engineering Applications of Artificial Intelligence}{126}{}{,}
\newblock
\begin{APACrefDOI} \doi{10.1016/j.engappai.2023.107072} \end{APACrefDOI}
\newblock

\newblock

\PrintBackRefs{\CurrentBib}

\bibitem [\protect \citeauthoryear {%
J.~Xu%
, Wu%
, Wang%
\BCBL {}\ \BBA {} Long%
}{%
J.~Xu%
\ \protect \BOthers {.}}{%
{\protect \APACyear {2022}}%
}]{%
Xu2022}
\APACinsertmetastar {%
Xu2022}%
\begin{APACrefauthors}%
Xu, J.%
, Wu, H.%
, Wang, J.%
\BCBL {} Long, M.%
\end{APACrefauthors}%
\unskip\
\newblock
\APACrefYearMonthDay{2022}{6}{}.
\newblock
{\BBOQ}\APACrefatitle {{Anomaly Transformer: Time Series Anomaly Detection with Association Discrepancy}} {{Anomaly Transformer: Time Series Anomaly Detection with Association Discrepancy}}.{\BBCQ}
\newblock
\APACjournalVolNumPages{arXiv}{}{}{,}
\newblock

\newblock

\PrintBackRefs{\CurrentBib}

\bibitem [\protect \citeauthoryear {%
R.~Xu%
, Song%
, Wu%
, Wang%
\BCBL {}\ \BBA {} Zhou%
}{%
R.~Xu%
\ \protect \BOthers {.}}{%
{\protect \APACyear {2025}}%
}]{%
Xu2025Change-pointReview}
\APACinsertmetastar {%
Xu2025Change-pointReview}%
\begin{APACrefauthors}%
Xu, R.%
, Song, Z.%
, Wu, J.%
, Wang, C.%
\BCBL {} Zhou, S.%
\end{APACrefauthors}%
\unskip\
\newblock
\APACrefYearMonthDay{2025}{3}{}.
\newblock
\APACrefbtitle {{Change-point detection with deep learning: A review}} {{Change-point detection with deep learning: A review}}\ (\BVOL~12)\ (\BNUM~1).
\newblock
\APACaddressPublisher{}{Higher Education Press Limited Company}.
\PrintBackRefs{\CurrentBib}

\bibitem [\protect \citeauthoryear {%
Yang%
\ \protect \BOthers {.}}{%
Yang%
\ \protect \BOthers {.}}{%
{\protect \APACyear {2022}}%
}]{%
Yang2022ADetection}
\APACinsertmetastar {%
Yang2022ADetection}%
\begin{APACrefauthors}%
Yang, Z.%
, Liu, X.%
, Li, T.%
, Wu, D.%
, Wang, J.%
, Zhao, Y.%
\BCBL {} Han, H.%
\end{APACrefauthors}%
\unskip\
\newblock
\APACrefYearMonthDay{2022}{5}{}.
\newblock
{\BBOQ}\APACrefatitle {{A systematic literature review of methods and datasets for anomaly-based network intrusion detection}} {{A systematic literature review of methods and datasets for anomaly-based network intrusion detection}}.{\BBCQ}
\newblock
\APACjournalVolNumPages{Computers {\&} Security}{116}{}{102675,}
\newblock
\begin{APACrefDOI} \doi{10.1016/j.cose.2022.102675} \end{APACrefDOI}
\newblock

\newblock

\PrintBackRefs{\CurrentBib}

\bibitem [\protect \citeauthoryear {%
Ye%
\ \BBA {} Chen%
}{%
Ye%
\ \BBA {} Chen%
}{%
{\protect \APACyear {2001}}%
}]{%
Ye2001AnSystems}
\APACinsertmetastar {%
Ye2001AnSystems}%
\begin{APACrefauthors}%
Ye, N.%
\BCBT {}\ \BBA {} Chen, Q.%
\end{APACrefauthors}%
\unskip\
\newblock
\APACrefYearMonthDay{2001}{3}{}.
\newblock
{\BBOQ}\APACrefatitle {{An anomaly detection technique based on a chi‐square statistic for detecting intrusions into information systems}} {{An anomaly detection technique based on a chi‐square statistic for detecting intrusions into information systems}}.{\BBCQ}
\newblock
\APACjournalVolNumPages{Quality and Reliability Engineering International}{17}{2}{105--112,}
\newblock
\begin{APACrefDOI} \doi{10.1002/qre.392} \end{APACrefDOI}
\newblock

\newblock

\PrintBackRefs{\CurrentBib}

\bibitem [\protect \citeauthoryear {%
Yu%
, Zhu%
, Li%
\BCBL {}\ \BBA {} Wan%
}{%
Yu%
\ \protect \BOthers {.}}{%
{\protect \APACyear {2014}}%
}]{%
Yu2014TimePrediction}
\APACinsertmetastar {%
Yu2014TimePrediction}%
\begin{APACrefauthors}%
Yu, Y.%
, Zhu, Y.%
, Li, S.%
\BCBL {} Wan, D.%
\end{APACrefauthors}%
\unskip\
\newblock
\APACrefYearMonthDay{2014}{}{}.
\newblock
{\BBOQ}\APACrefatitle {{Time series outlier detection based on sliding window prediction}} {{Time series outlier detection based on sliding window prediction}}.{\BBCQ}
\newblock
\APACjournalVolNumPages{Mathematical Problems in Engineering}{2014}{}{,}
\newblock
\begin{APACrefDOI} \doi{10.1155/2014/879736} \end{APACrefDOI}
\newblock

\newblock

\PrintBackRefs{\CurrentBib}

\bibitem [\protect \citeauthoryear {%
R.~Zhang%
\ \protect \BOthers {.}}{%
R.~Zhang%
\ \protect \BOthers {.}}{%
{\protect \APACyear {2020}}%
}]{%
Zhang2020Correlation-AwareNetworks}
\APACinsertmetastar {%
Zhang2020Correlation-AwareNetworks}%
\begin{APACrefauthors}%
Zhang, R.%
, Hao, Y.%
, Yu, D.%
, Chang, W\BHBI C.%
, Lai, G.%
\BCBL {} Yang, Y.%
\end{APACrefauthors}%
\unskip\
\newblock
\APACrefYearMonthDay{2020}{}{}.
\newblock
{\BBOQ}\APACrefatitle {{Correlation-Aware Change-Point Detection via Graph Neural Networks}} {{Correlation-Aware Change-Point Detection via Graph Neural Networks}}.{\BBCQ}
\newblock
 \APACrefbtitle {Lecture Notes in Computer Science (including subseries Lecture Notes in Artificial Intelligence and Lecture Notes in Bioinformatics)} {Lecture notes in computer science (including subseries lecture notes in artificial intelligence and lecture notes in bioinformatics)}\ (\BVOL\ 12534 LNCS, \BPGS\ 555--567).
\newblock
\APACaddressPublisher{}{Springer Science and Business Media Deutschland GmbH}.
\PrintBackRefs{\CurrentBib}

\bibitem [\protect \citeauthoryear {%
Y.~Zhang%
, An%
\BCBL {}\ \BBA {} Shen%
}{%
Y.~Zhang%
\ \protect \BOthers {.}}{%
{\protect \APACyear {2017}}%
}]{%
Zhang2017FastSpectroscopy}
\APACinsertmetastar {%
Zhang2017FastSpectroscopy}%
\begin{APACrefauthors}%
Zhang, Y.%
, An, L.%
\BCBL {} Shen, J.%
\end{APACrefauthors}%
\unskip\
\newblock
\APACrefYearMonthDay{2017}{8}{}.
\newblock
{\BBOQ}\APACrefatitle {{Fast computation of full density matrix of multispin systems for spatially localized in vivo magnetic resonance spectroscopy}} {{Fast computation of full density matrix of multispin systems for spatially localized in vivo magnetic resonance spectroscopy}}.{\BBCQ}
\newblock
\APACjournalVolNumPages{Medical Physics}{44}{8}{4169--4178,}
\newblock
\begin{APACrefDOI} \doi{10.1002/mp.12375} \end{APACrefDOI}
\newblock

\newblock

\PrintBackRefs{\CurrentBib}

\bibitem [\protect \citeauthoryear {%
Zhao%
\ \protect \BOthers {.}}{%
Zhao%
\ \protect \BOthers {.}}{%
{\protect \APACyear {2020}}%
}]{%
Zhao2020MultivariateNetwork}
\APACinsertmetastar {%
Zhao2020MultivariateNetwork}%
\begin{APACrefauthors}%
Zhao, H.%
, Wang, Y.%
, Duan, J.%
, Huang, C.%
, Cao, D.%
, Tong, Y.%
\BDBL {}Zhang, Q.%
\end{APACrefauthors}%
\unskip\
\newblock
\APACrefYearMonthDay{2020}{11}{}.
\newblock
{\BBOQ}\APACrefatitle {{Multivariate time-series anomaly detection via graph attention network}} {{Multivariate time-series anomaly detection via graph attention network}}.{\BBCQ}
\newblock
 \APACrefbtitle {Proceedings - IEEE International Conference on Data Mining, ICDM} {Proceedings - ieee international conference on data mining, icdm}\ (\BVOL\ 2020-November, \BPGS\ 841--850).
\newblock
\APACaddressPublisher{}{Institute of Electrical and Electronics Engineers Inc.}
\PrintBackRefs{\CurrentBib}

\bibitem [\protect \citeauthoryear {%
Zhong%
\ \protect \BOthers {.}}{%
Zhong%
\ \protect \BOthers {.}}{%
{\protect \APACyear {2025}}%
}]{%
Zhong2025PatchAD:Detection}
\APACinsertmetastar {%
Zhong2025PatchAD:Detection}%
\begin{APACrefauthors}%
Zhong, Z.%
, Yu, Z.%
, Yang, Y.%
, Wang, W.%
, Yang, K.%
\BCBL {} Chen, C.L.%
\end{APACrefauthors}%
\unskip\
\newblock
\APACrefYearMonthDay{2025}{}{}.
\newblock
{\BBOQ}\APACrefatitle {{PatchAD: A Lightweight Patch-Based MLP-Mixer for Time Series Anomaly Detection}} {{PatchAD: A Lightweight Patch-Based MLP-Mixer for Time Series Anomaly Detection}}.{\BBCQ}
\newblock
\APACjournalVolNumPages{IEEE Transactions on Big Data}{}{}{,}
\newblock
\begin{APACrefDOI} \doi{10.1109/TBDATA.2025.3596745} \end{APACrefDOI}
\newblock

\newblock

\PrintBackRefs{\CurrentBib}

\end{thebibliography}

\appendix
\section{Architecture Details} \label{appendix:archdetails}
This appendix provides detailed architectural configurations for each experiment described in the main text. These tables include layer types, shapes, and parameter settings used in the models.

\begin{table}[h!]
\centering
\caption{Architecture details for proportionality test (Sec.~\ref{section:propexp})}
\label{tab:prop_test_sec_41}
\begin{tabular}{@{}lll@{}}
\toprule
\textbf{Component} & \textbf{Parameter} & \textbf{Value} \\
\midrule
\multirow{5}{*}{\textbf{Parameter}} & Sequence Length (L) & 1 \\
 & Past Steps ($N_p$) & 1 \\
 & Forecast Steps ($N_f$) & 1 \\
 & Latent Size ($N_e$) & 4 \\
\midrule
\multirow{5}{*}{\textbf{Encoder (E)}} & Input Shape & (L,) \\
 & Layer Types & Dense \\
 & Architecture & 1 Dense Layer \\
 & Units/Filters & 4 (Linear) \\
 & Output Shape & ($N_e$,) \\
\midrule
\multirow{5}{*}{\textbf{Recurrent (G)}} & Input Shape & ($N_p, N_e$) \\
 & Layer Type & GRU \\
 & Architecture & 1 GRU Layer \\
 & Units & $N_g = 8$ \\
 & Output Shape & ($N_p, N_g$) \\
\midrule
\multirow{6}{*}{\textbf{Forecasting (F$_i$)}} & Input Shape & ($N_g$,) \\
 & Layer Types & Dense, ReLU, Linear, Exponential \\
 & Architecture & 6 Dense (ReLU) + 2 Output Dense \\
 & Units (Dense Layers) & [16, 16, 32, 32, 64, 64] \\
 & Units (Output Layers) & $N_e$ (Linear mean), $N_e$ (Exp std) \\
 & Output Shape & ($N_e$, 2) \\
\midrule
\multirow{5}{*}{\textbf{Decoder (D)}} & Input Shape & ($N_e$,) \\
 & Layer Types & Dense, Linear \\
 & Architecture & 1 Dense Layer \\
 & Units/Filters & 1 (Linear) \\
 & Output Shape & (L,) \\
\bottomrule
\end{tabular}
\end{table}

\begin{table}[h!]
\centering
\caption{Architecture details for sine wave experiment (Sec. \ref{section:sineexp})}
\label{tab:sine_wave_sec_42}
\begin{tabular}{@{}lll@{}}
\toprule
\textbf{Component} & \textbf{Parameter} & \textbf{Value} \\
\midrule
\multirow{5}{*}{\textbf{Parameter}} & Sequence Length (L) & 256 \\
 & Past Steps ($N_p$) & 5 \\
 & Forecast Steps ($N_f$) & 3 \\
 & Latent Size ($N_e$) & 16 \\
\midrule
\multirow{5}{*}{\textbf{Encoder (E)}} & Input Shape & (L, 1) \\
 & Layer Types & Conv, ReLU, BN, MaxPool, Dense \\
 & Architecture & 2 Conv Blocks + Dense \\
 & Units/Filters & 32, 64 (Conv) + 16 (Dense) \\
 & Output Shape & ($N_e$,) \\
\midrule
\multirow{5}{*}{\textbf{Recurrent (G)}} & Input Shape & ($N_p, N_e$) \\
 & Layer Type & GRU \\
 & Architecture & 1 GRU Layer \\
 & Units & $N_g = 32$ \\
 & Output Shape & ($N_p, N_g$) \\
\midrule
\multirow{6}{*}{\textbf{Forecasting (F$_i$)}} & Input Shape & ($N_g$,) \\
 & Layer Types & Dense, ReLU, Linear, Exponential \\
 & Architecture & 3 Dense (ReLU) + 2 Output Dense \\
 & Units (Dense Layers) & [64, 128, 256] \\
 & Units (Output Layers) & $N_e$ (Linear mean), $N_e$ (Exp std) \\
 & Output Shape & ($N_e$, 2) \\
\midrule
\multirow{5}{*}{\textbf{Decoder (D)}} & Input Shape & ($N_e$,) \\
 & Layer Types & Dense, ConvTransp/Upsample, ReLU, Linear \\
 & Architecture & Dense + Reshape + 2 Conv Blocks + Conv \\
 & Units/Filters & 64, 32 (Conv/ConvTransp) \\
 & Output Shape & (L, 1) \\
\bottomrule
\end{tabular}
\end{table}

\begin{table}[h!]
\centering
\caption{Architecture details for counting with MNIST (Sec. \ref{section:mnistexp})}
\label{tab:mnist_sec_43}
\begin{tabular}{@{}lll@{}}
\toprule
\textbf{Component} & \textbf{Parameter} & \textbf{Value} \\
\midrule
\multirow{5}{*}{\textbf{Parameter}} & Input size & (28, 28, 1) \\
 & Past Steps ($N_p$) & 1 \\
 & Forecast Steps ($N_f$) & 1 \\
 & Latent Size ($N_e$) & 16 \\
\midrule
\multirow{5}{*}{\textbf{Encoder (E)}} & Input Shape & (28, 28, 1) \\
 & Layer Types & Conv, ReLU, MaxPool, Flatten, Dense \\
 & Architecture & 2 Conv Blocks + Flatten + Dense \\
 & Units/Filters & 32, 64 (Conv) + 16 (Dense) \\
 & Output Shape & ($N_e$,) \\
\midrule
\multirow{5}{*}{\textbf{Recurrent (G)}} & Input Shape & ($N_p, N_e$) \\
 & Layer Type & GRU \\
 & Architecture & 1 GRU Layer \\
 & Units & $N_g = 32$ \\
 & Output Shape & ($N_p, N_g$) \\
\midrule
\multirow{6}{*}{\textbf{Forecasting (F$_i$)}} & Input Shape & ($N_g$,) \\
 & Layer Types & Dense, ReLU, Linear, Exponential \\
 & Architecture & 3 Dense (ReLU) + 2 Output Dense \\
 & Units (Dense Layers) & [64, 128, 256] \\
 & Units (Output Layers) & $N_e$ (Linear mean), $N_e$ (Exp std) \\
 & Output Shape & ($N_e$, 2) \\
\midrule
\multirow{5}{*}{\textbf{Decoder (D)}} & Input Shape & ($N_e$,) \\
 & Layer Types & Dense, Reshape, ConvTransp/Upsample, ReLU, Sigmoid \\
 & Architecture & Dense + Reshape + 2 Conv Blocks + Conv \\
 & Units/Filters & 64, 32 (Conv/ConvTransp) + 1 (Conv) \\
 & Output Shape & (28, 28, 1) \\
\bottomrule
\end{tabular}
\end{table}

\begin{table}[h!]
\centering
\caption{Architecture details for the artifact detection in MRSI (Sec. \ref{section:mrsiexp})}
\label{tab:mrsi_sec_44}
\begin{tabular}{@{}lll@{}}
\toprule
\textbf{Component} & \textbf{Parameter} & \textbf{Value} \\
\midrule
\multirow{5}{*}{\textbf{Parameter}} & Sequence Length (L) & 32 \\
 & Past Steps ($N_p$) & 1 \\
 & Forecast Steps ($N_f$) & 1 \\
 & Latent Size ($N_e$) & 16 \\
\midrule
\multirow{5}{*}{\textbf{Encoder (E)}} & Input Shape & (L,) \\
 & Layer Types & Dense, ReLU, Linear \\
 & Architecture & 4 Dense (ReLU) + 1 Dense (Linear) \\
 & Units/Filters & 512 (Dense), 16 (Dense) \\
 & Output Shape & ($N_e$,) \\
\midrule
\multirow{5}{*}{\textbf{Recurrent (G)}} & Input Shape & ($N_p, N_e$) \\
 & Layer Type & GRU \\
 & Architecture & 1 GRU Layer \\
 & Units & $N_g = 32$ \\
 & Output Shape & ($N_p, N_g$) \\
\midrule
\multirow{6}{*}{\textbf{Forecasting (F$_i$)}} & Input Shape & ($N_g$,) \\
 & Layer Types & Dense, ReLU, Linear, Exponential \\
 & Architecture & 6 Dense (ReLU) + 2 Output Dense \\
 & Units (Dense Layers) & [64, 64, 128, 128, 256, 256] \\
 & Units (Output Layers) & $N_e$ (Linear mean), $N_e$ (Exp std) \\
 & Output Shape & ($N_e$, 2) \\
\midrule
\multirow{5}{*}{\textbf{Decoder (D)}} & Input Shape & ($N_e$,) \\
 & Layer Types & Dense, ReLU, Linear \\
 & Architecture & 4 Dense (ReLU) + 1 Dense (Linear) \\
 & Units/Filters & 512 (Dense), L (Dense) \\
 & Output Shape & (L,) \\
\bottomrule
\end{tabular}
\end{table}

\end{document}